# FID-Net: A Feature-Enhanced Deep Learning Network for Forest Infestation Detection

Yan Zhang[a], Baoxin Li[a], Han Sun[b,c], Yuhang Gao[a], Mingtai Zhang[a], Pei Wang[a,*]

[a] School of Science, Beijing Forestry University, Beijing, 100083, China

[b] State Key Laboratory of Efficient Production of Forest Resources, Beijing Forestry University, Beijing, 100083, China

[c] School of Ecology and Nature Conservation, Beijing Forestry University, Beijing, 100083, China

[*] Corresponding author. E-mail address: wangpei@bjfu.edu.cn.

Abstract: Forest pests and diseases pose critical threats to the health and stability of forest ecosystems, thus demanding efficient and precise monitoring and analysis technologies. Addressing the limitations of traditional methods in large-scale, fine-grained detection of infected trees, this study focuses on the accurate identification of pest-affected trees and comprehensive analysis of infestation dynamics. We propose FID-Net, a deep learning-based detection model designed to recognize infected trees caused by pest infection from UAV visible-light imagery and conducts pest situation analysis through three metrics based on detection results. Built upon the classic YOLOv8n model, FID-Net innovatively proposes a lightweight Feature Enhancement Module (FEM), an Adaptive Multi-scale Feature Fusion Module (AMFM), and introduces an Efficient Channel Attention (ECA) mechanism. The FEM explicitly extracts disease-sensitive cues from RGB images, while the AMFM achieves semantic alignment and dynamic fusion of dual-branch features (original RGB and FEM-enhanced features). The ECA mechanism strengthens key discriminative information with minimal computational overhead. Based on the detection results, a comprehensive pest situation analysis framework was innovatively constructed by integrating three spatial analysis methods: (1) Kernel Density Estimation (KDE) to reveal spatial clustering hotspots of infected trees; (2) neighborhood evaluation to assess potential infection risks of healthy trees; (3) DBSCAN clustering to identify high-density healthy tree clusters for delineating priority protection areas. Experiments were conducted using UAV visible-light imagery collected from 32 forest sample plots in the eastern Tianshan region of Xinjiang, China. Results showed that FID-Net achieved 86.10% precision, 75.44% recall, 82.29% mAP@0.5, and 64.30% mAP@0.5:0.95 on the test set, outperforming mainstream YOLO series models. Situation analysis confirms significant clustering characteristics of infected trees, providing a scientific basis for prioritizing potential forest protection zones. This research demonstrates that FID-Net can accurately distinguish between infected and healthy trees, and when combined with spatial analysis metrics, can provide solid data support for intelligent monitoring, risk warning, and precise management of forest pests and diseases.

# 1. Introduction

Forests, as core components of the Earth's ecosystem, perform irreplaceable critical functions such as maintaining ecological balance, regulating global climate, conserving water resources, and protecting biodiversity(Harris et al., 2021; Pan et al., 2011). As the largest carbon reservoir in terrestrial ecosystems, forests play a strategic role in mitigating climate change and ensuring ecological security, while also supporting the achievement of global sustainable development goals(Calvin et al., 2023; Canadell and Raupach, 2008). However, forest pests and diseases, as one of the most destructive biological disasters, have emerged as primary threats to global forest health.

Against the backdrop of global climate change and intensified forest fragmentation, the frequency, spread, and severity of forest pest and disease outbreaks have increased significantly. Pine wilt disease has severely damaged pine forests across multiple countries, and the United States is projected to lose over 20% of basal area in approximately 25 million hectares of forest land due to pests and diseases by 2027("FAO," 2025; "FAO," 2024). The continuous spread of pests and diseases not only causes tree growth decline and mortality, disrupting forest structure and ecological integrity, but may also trigger cascading ecological effects such as reduced carbon sequestration capacity, habitat destruction, and intensified soil erosion(Liu et al., 2020; Yang et al., 2025), posing severe challenges to regional and global ecological security.

Therefore, timely and precise monitoring of forest pests and diseases is fundamental to understanding disaster dynamics, analyzing spread patterns, and scientifically implementing prevention and control measures. It provides reliable data support for risk assessment and outbreak early warning, helps identify priority areas and critical periods for intervention, and avoids resource waste and environmental impacts from blind prevention, thus maximizing disaster control efficiency and safeguarding forest resources and ecosystem stability.

Traditional forest pest and disease survey methods primarily include ground-based field surveys, fixed-plot monitoring, aerial visual inspections, and satellite remote sensing(Brovkina et al., 2018; Dash et al., 2017; Lausch et al., 2018; Zhang et al., 2021). Large-scale area investigations predominantly rely on satellite remote sensing and aerial inspections: satellite remote sensing, while capable of wide coverage, is constrained by spatial and temporal resolution limitations, making it difficult to capture early-stage or small-scale pest symptoms and exhibiting weak recognition capability for understory pest conditions; aerial visual inspections depend heavily on manual expertise, suffering from subjectivity, low efficiency, and high costs, rendering them unsuitable for routine monitoring of complex terrains and large forest areas. Ground-based surveys can obtain detailed pest information (such as disease types and pest density), but are time-consuming, labor-intensive, and limited in coverage, failing to meet the needs of rapid large-scale monitoring and facing significant implementation difficulties in remote or steep areas.

With the growing demand for high-precision and efficient monitoring in forest ecological protection, there is an urgent need to develop monitoring technologies that combine wide-area coverage with fine-grained detection capabilities. Unmanned Aerial Vehicle (UAV) remote sensing technology, characterized by flexibility, simple operation, and controllable costs, can be equipped with high-resolution optical cameras and multispectral sensors to rapidly acquire high-spatial-resolution imagery of forest canopies. This enables precise identification of subtle early symptoms caused by pests and diseases, such as leaf discoloration and wilting(Deng et al., 2020; Huo et al., 2023; Junttila et al., 2022). Meanwhile, UAVs can flexibly integrate high-frequency small-areas monitoring with rapid surveys, effectively compensating for the limitations of satellite remote sensing and ground surveys, thereby providing a novel solution for refined and routine forest pest monitoring.

UAV-based tree pest monitoring can be categorized into three main types. The first type comprises early studies based on optical imagery, multispectral, or hyperspectral data, generally employing traditional machine learning models such as Random Forest and Support Vector Machines for infected tree identification. Since these algorithms rely on handcrafted features, most studies use multispectral or hyperspectral data with rich spectral information combined with machine learning algorithms (Iordache et al., 2020; Wu et al., 2023; Yu et al., 2022; Zhang et al., 2018), while visible-light imagery—with limited information—has been rarely used (Hu et al., 2020; Oide et al., 2022).

The second type focuses on integrating multi-source data with UAV data to improve classification accuracy of infected trees. For instance, Qin et al. stacked visible-light and six-band multispectral imagery, applying an improved YOLOv5 model to achieve high-precision detection in complex terrains(Qin et al., 2023); Yu et al. combined hyperspectral data with LiDAR canopy structure to increase five-stage infection classification accuracy from 66.9% to 74%(Yu et al., 2021); Oblinger et al.

fused hyperspectral imagery with LiDAR data to implement four-level individual tree health classification(Oblinger et al., 2022). These studies demonstrate that coupling hyperspectral, visible-light, and LiDAR information on UAV platforms with feature fusion strategies using deep learning or machine learning can enhance model recognition capabilities for infected trees. However, these methods require multiple sensors, resulting in high costs and complex data processing.

The third type employs deep learning frameworks for automatic and precise infected tree identification from UAV remote sensing images. These methods can be further divided into single-stage (e.g., YOLO series) and two-stage (e.g., Faster R-CNN) models. Single-stage models are more efficient on the YOLO framework (S. Wang et al., 2023; Ye et al., 2024; Zhang et al., n.d.; Zhu et al., 2024). Early studies achieved rapid localization of infected pines using only visible-light images (Hu et al., 2022a; Li et al., 2021; Ye et al., 2024; Zhou et al., 2022); subsequent improvements enhanced detection through attention mechanism optimization or Transformer architecture integration(Chen et al., 2025; Dong et al., 2024; Li et al., 2024; Yuan et al., 2024). Two-stage models, such as Faster R-CNN, can also implement forest pest detection(Hu et al., 2022b; Wu and Jiang, 2023), but suffer from weaker generalization capability, high computational costs, and slow inference speeds, limiting their deployment for real-time large-scale forest patrols.

However, all three types of methods have limitations. The first two are constrained by high costs, operational complexity, and environmental interference associated with multispectral/hyperspectral/LiDAR equipment. And the third type methods face challenges including sparse infected tree distributions in application scenarios and insufficient validation in large-scale infection situations. Furthermore, simple infected tree detection alone cannot fully describe forest health status, analyze pest dynamics, predict spread risks, or support precise prevention and control.

Therefore, this study proposes FID-Net (A Feature-Enhanced Deep Learning Network for Forest Infestation Detection), a UAV remote sensing image detection model for infected trees, and conducts infected tree identification experiments in the eastern Tianshan Mountains of Xinjiang, China. Built upon YOLOv8n, this model innovatively constructs a lightweight Feature Enhancement Module (FEM) to generate task-oriented visible-light pest-sensitive features, an Adaptive Multi-scale Feature Fusion Module (AMFM) to achieve semantic alignment and dynamic fusion of dual-branch features, and embeds an Efficient Channel Attention (ECA) mechanism to strengthen key discriminative information. We also establish a high-quality UAV detection dataset for eastern Tianshan pine forests (East Tianshan Tree Detection Dataset, ET-TDD). Based on detection results, Kernel Density Estimation (KDE), neighborhood risk assessment, and Density-Based Spatial Clustering of Applications with Noise (DBSCAN) were integrated to develop a comprehensive forest pest situation analysis and risk early-warning methodology, providing technical support for precise monitoring and intelligent prevention and control of forest pests and diseases.

## 2. Materials and Methods

## 2.1 Study Area

The study area is situated in the eastern Tianshan region within Hami City, Xinjiang Uygur Autonomous Region, China (92°30′–94°10′E, 42°20′–43°50′N), covering a total area of approximately 3,200 km². Specifically located in the eastern segment of the Tianshan Mountain range (Fig. 1), the region features a typical temperate continental arid climate, with a mean annual temperature of ~3.5 °C and annual precipitation ranging from 250 to 450 mm—primarily concentrated in spring and early summer. Topographically, the terrain slopes upward gently from west to east, with elevations ranging from 1,800 to 3,200 m. Bordered to the south by the main ridge of the eastern Tianshan Mountains and to the north by the desert of Hami Basin, the area forms distinct vertical climatic zones. Mountainous coniferous forests cover approximately 680 km² of the region, dominated by Siberian larch (*Larix sibirica*) and Schrenk spruce (*Picea schrenkiana*). These forests are primarily distributed across shaded slopes, semi-shaded slopes, and valley floors, serving as a key component of the eastern Tianshan forest ecosystem.

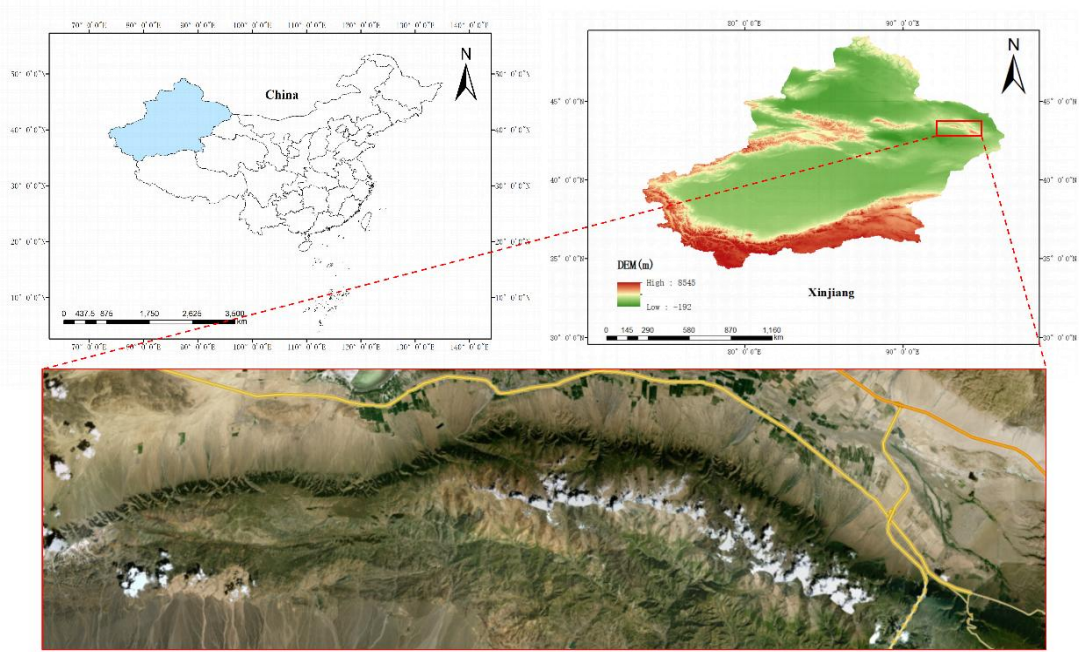

**Fig. 1.** Study area

In recent years, the eastern Tianshan Mountains—an important ecological barrier in northwestern China—have suffered forest degradation caused by bark beetle infestations. These pests spread rapidly via short-distance flight and infested timber transportation. Their feeding behavior, coupled with infections by symbiotic fungi, induces phloem necrosis and sapwood blue-stain disease, ultimately leading to needle discoloration, diminished resin secretion, and eventual whole-tree death(Netherer et al., 2024; Singh et al., 2024). Infected trees typically exhibit subtle early symptoms and deteriorate rapidly, culminating in large-scale forest die-off over a short timeframe and posing a significant threat to regional ecological security.

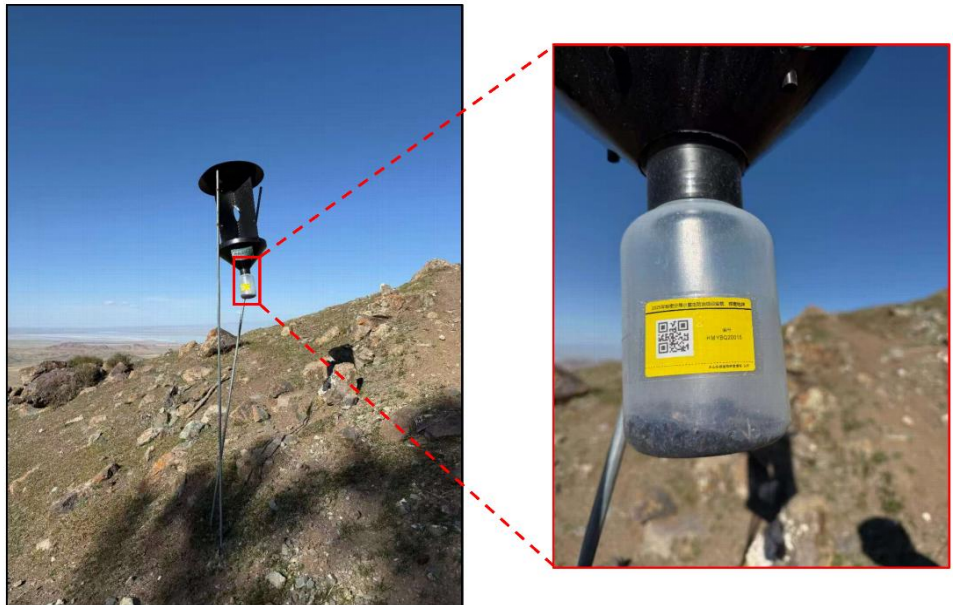

(a)

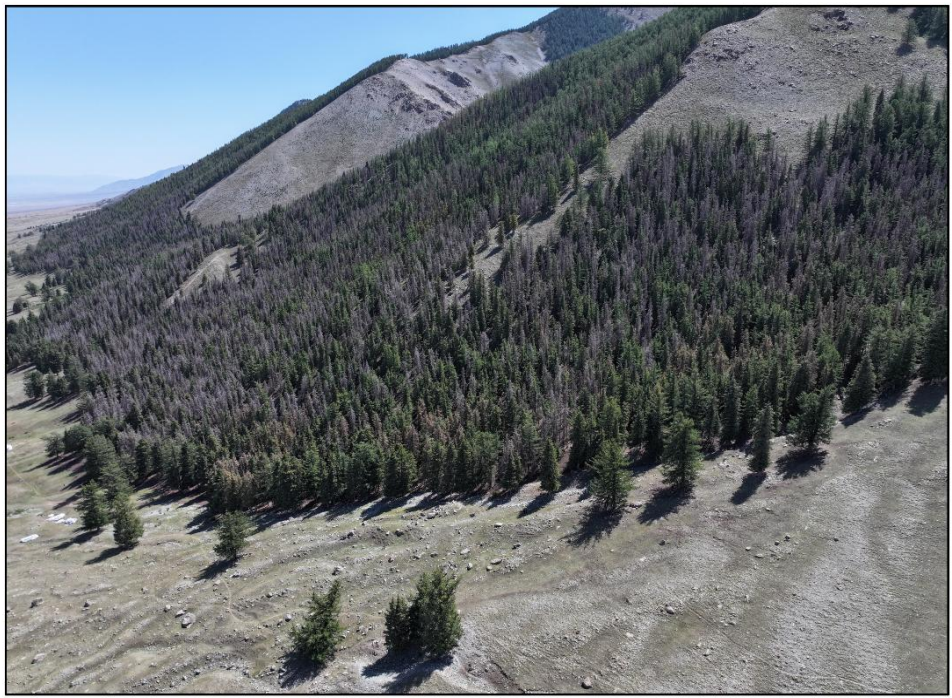

(b)

**Fig. 2.** (a) Local bark beetle trapping device; (b) Forest damage conditions in the study area.

Typical indicators of bark beetle outbreaks have already appeared in local forests. As illustrated in **Fig. 2(a)**, considerable numbers of adult beetles have been captured by devices deployed by forestry departments, indicating high population density and active beetle movement. Meanwhile, **Fig. 2(b)** shows substantial forest damage, with extensive patches of dead trees forming contiguous mortality zones, confirming the significant ecological impacts of the infestation.

## 2.2 Data Collection

From 16 to 22 August 2025, 32 sample plots along the east–west axis of the Tianshan Mountains were established, covering nearly 1,500 km, and UAV-based data collection was conducted at each plot. Field observations revealed substantially higher tree mortality rate in the western section compared with the eastern section; thus, the sampling density and flight frequency were increased in the west, while sample plots in the eastern section were spaced at wider intervals along the mountain range. The spatial distribution of sample plots is illustrated in Fig. 3, where the red boxes indicate planned sampling zones.

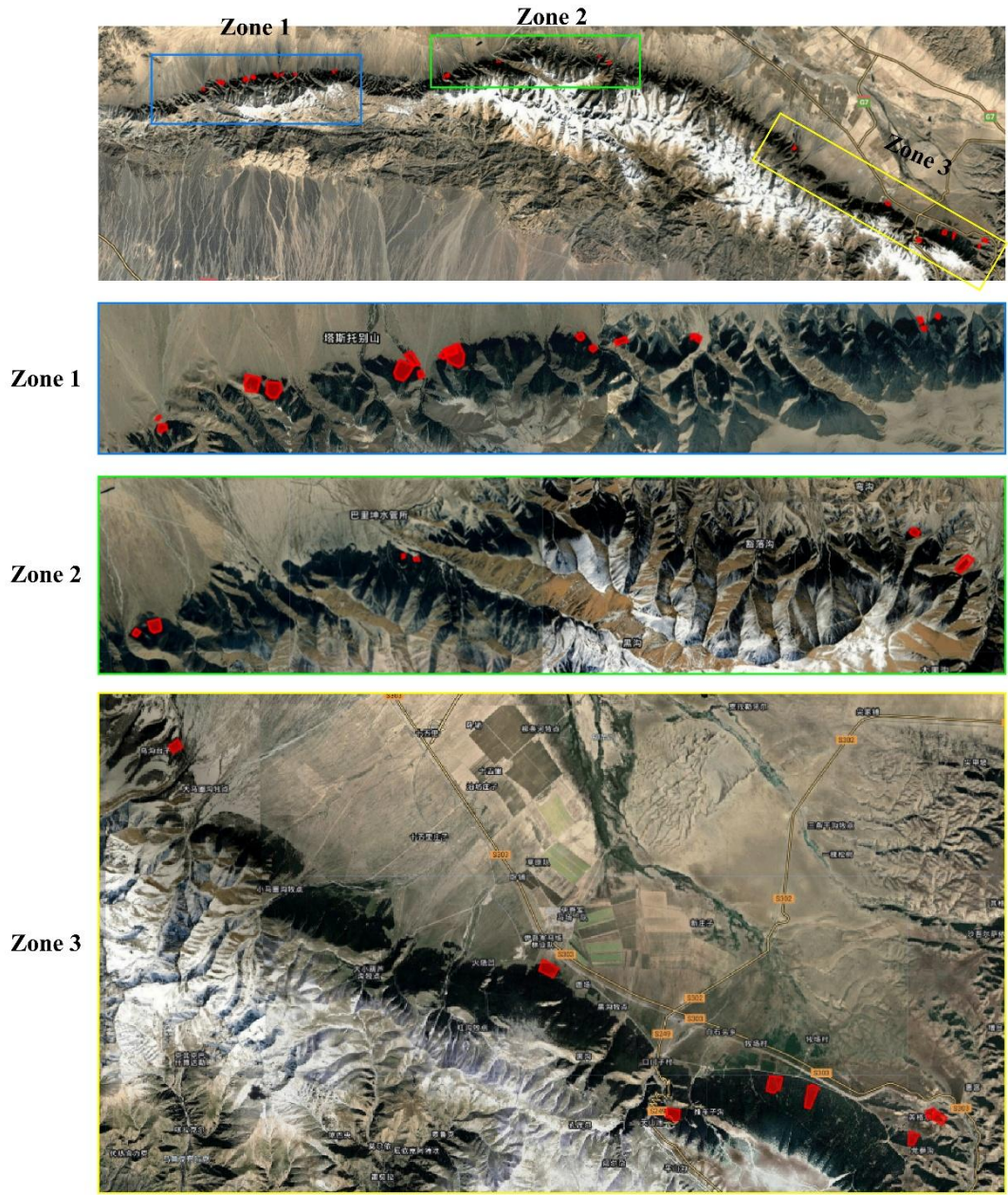

**Fig. 3.** Distribution of sample plots

A DJI Mavic 3 Pro UAV equipped with a Hasselblad camera was used to capture imagery at each sample plot. The camera features an effective pixel resolution of approximately 20 megapixels (MP) with a focal length of 24 mm (full-frame equivalent). Given that most trees in the study area are distributed across mid-slope regions with significant topographic variation, the UAV's flight altitude was dynamically adjusted based on local terrain conditions, ranging from 40 m to 500 m.

Image overlap rates were set to 70–80%. Acquired images were inspected for clarity, and suboptimal ones were excluded. Ultimately, a total of 3,967 high-quality UAV images were obtained. **Fig. 4** shows representative examples captured under different illumination scenarios (bright sunlight, overcast, cloudy) and at various flight altitudes.

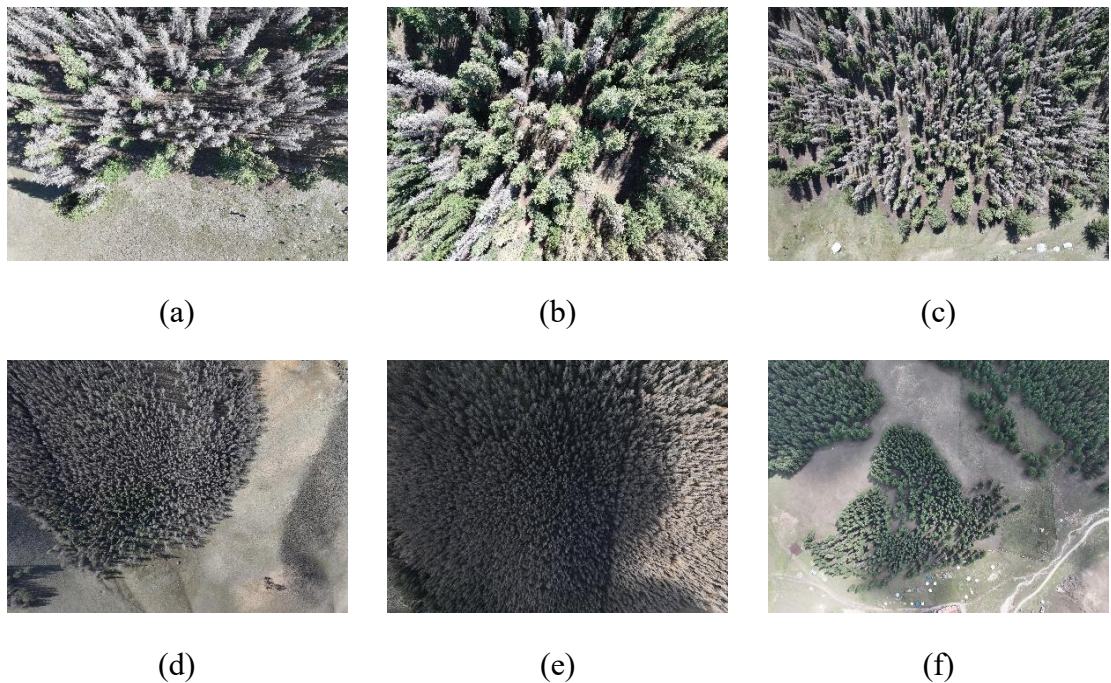

**Fig. 4.** Data examples. (a) 60 m, strong sunlight; (b) 80 m, strong sunlight; (c) 100 m, strong sunlight; (d) 420 m, evening; (e) 200 m, cloudy conditions; (f) 420 m, cloudy conditions.

## 2.3 Methods

Based on field survey and UAV-based data collection, this study developed the technical workflow illustrated in **Fig. 5** for detecting and analyzing infected trees in the eastern Tianshan Mountains. The proposed workflow consists of three major components. First, UAV images were orthorectified, cropped, and manually filtered to construct the ET-TDD. Second, YOLOv8n was selected as the baseline model, and a novel network framework named FID-Net was further developed by integrating a FEM, an AMFM and an ECA mechanism. This optimized framework enables the concurrent identification of both infected and healthy trees. Third, comprehensive performance evaluation and ablation experiments were conducted. In addition, KDE, neighborhood estimation and DBSCAN were employed to analyze spatial patterns of infected trees, thereby providing data-driven insights for precision pest management.

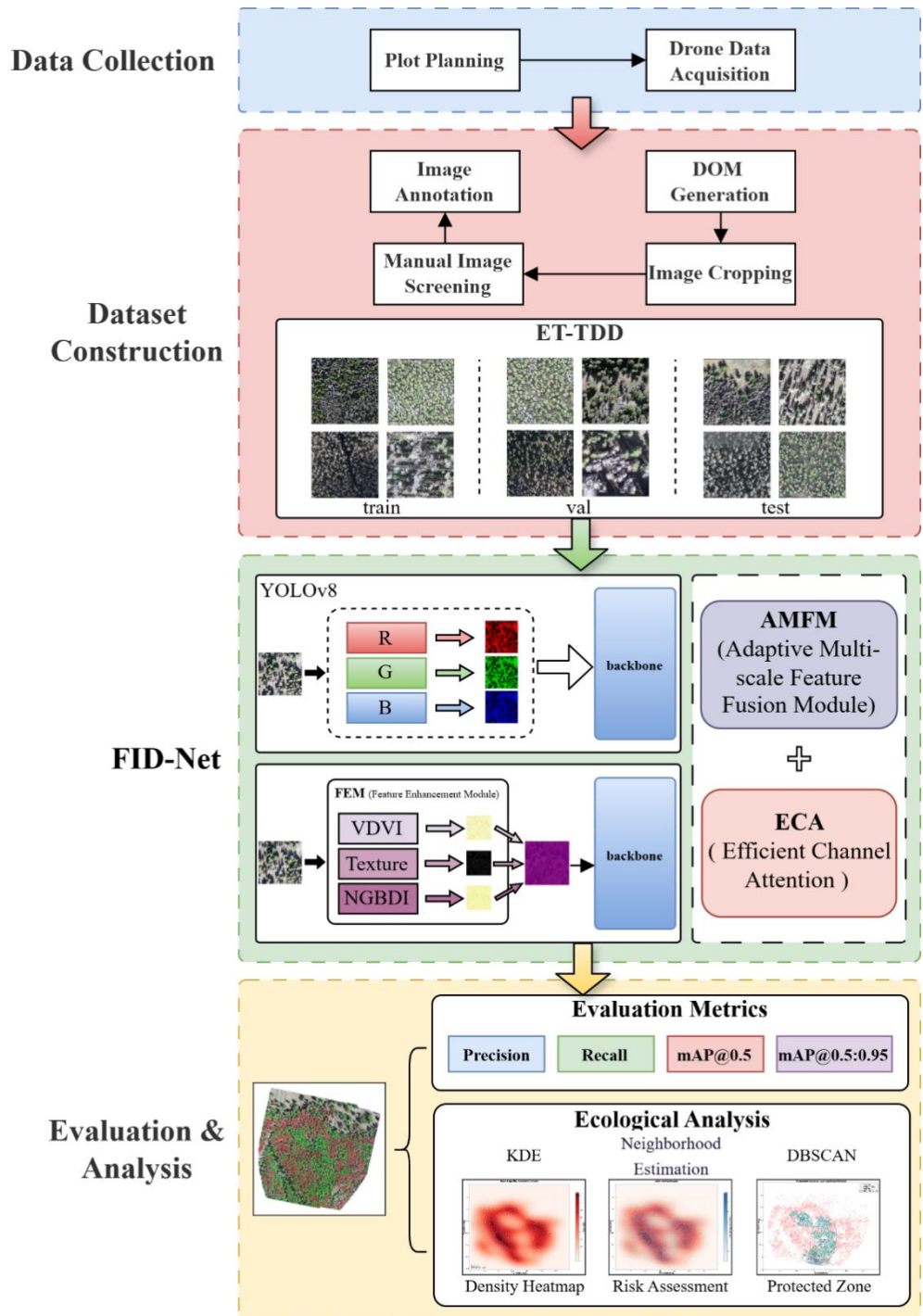

**Fig. 5.** Technical workflow

*2.3.1 Dataset Construction*

To balance training efficiency and fine-grained feature preservation, each Digital Orthophoto Map (DOM) was uniformly segmented into 1024 × 1024-pixel image tiles. However, flight-related artifacts (e.g., wind-induced motion blur, terrain undulation-derived distortion, and sensor-related aberrations) degraded the visual quality of partial tiles. Therefore, all cropped tiles underwent manual quality inspection, with low-quality images excluded, resulting in 312 high-quality images for subsequent analysis.

Subsequently, the LabelImg tool was utilized to manually annotate all tree instances within the retained images, generating a total of 40,241 bounding boxes. Based on this standardized workflow, we constructed the high-quality UAV-based tree detection dataset for the eastern Tianshan forests—the ET-TDD—which covers 32 sample plots and capturing diverse stand conditions with varying infestation severity and environmental contexts.

To evaluate the model's generalization capability, representative plots with severe, moderate, and mild infestation levels, as well as diverse illumination conditions (e.g., strong sunlight, cloudy skies), were included in the test set. The remaining plots were divided into training and validation sets at an 8:2 ratio, resulting in 234 training images and 42 validation images, ensuring robustness across diverse scenarios. Fig. 6 presents representative examples from Sample Plot 1 and Sample Plot 2.

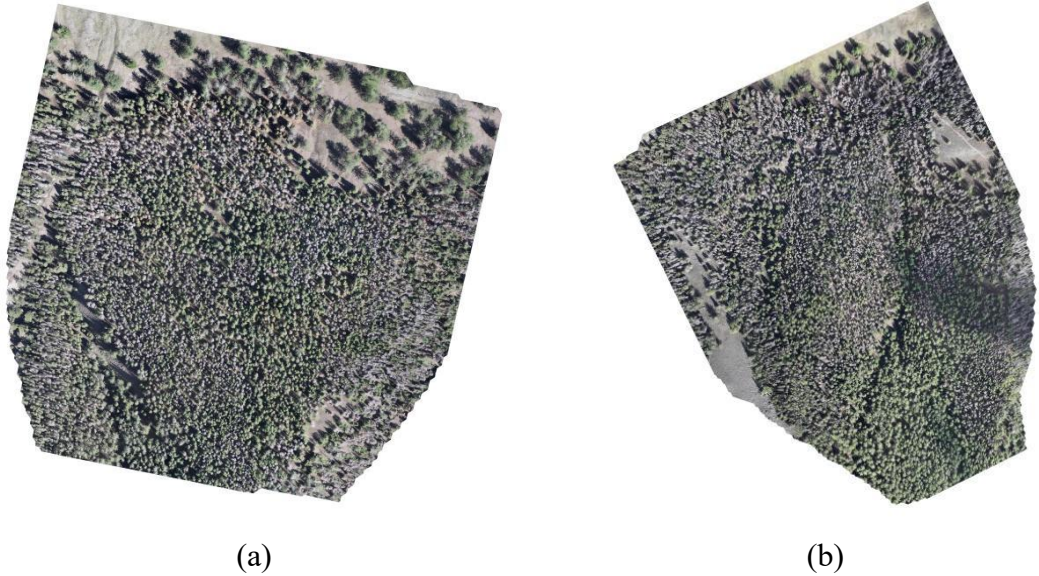

                    (a)                              (b)

**Fig. 6.** UAV-based orthomosaic of the study area. (a) Sample plot 1; (b) Sample plot 2.

*2.3.2 FID-Net*

In deep learning-based object detection, one-stage detectors are well suited for real-time pest and disease monitoring in remote sensing imagery due to their end-to-end structure and high inference efficiency. Among them, the YOLO family provides a favorable balance between accuracy and speed. After comprehensive evaluation, YOLOv8n was selected as the baseline model for this study and further optimized through the integration of the proposed modules to develop the FID-Net architecture.

YOLOv8 is a general-purpose object detection framework proposed by Ultralytics in 2023(Sapkota et al., 2024; Sharma et al., 2024; Solimani et al., 2024). The model consists of three major components: Backbone, Neck, and Head. The Backbone adopts an improved CSPDarknet (Cross Stage Partial Darknet) structure and introduces the C2f module (Cross Stage Partial with 2f), which enhances gradient propagation efficiency through branched feature flow while reducing computational redundancy and improving the robustness of feature extraction. The Neck integrates concepts from the Feature Pyramid Network (FPN) and Path Aggregation Network (PAN) to achieve efficient cross-scale feature fusion, enabling the model to simultaneously capture targets of different sizes and semantic levels. The Detection Head adopts an anchor-free design instead of the traditional anchor-based mechanism, predicting object centers and bounding box sizes through distributional regression, thereby significantly improving small-object detection accuracy and training stability. The overall architecture of YOLOv8 is shown in **Fig. 7**.

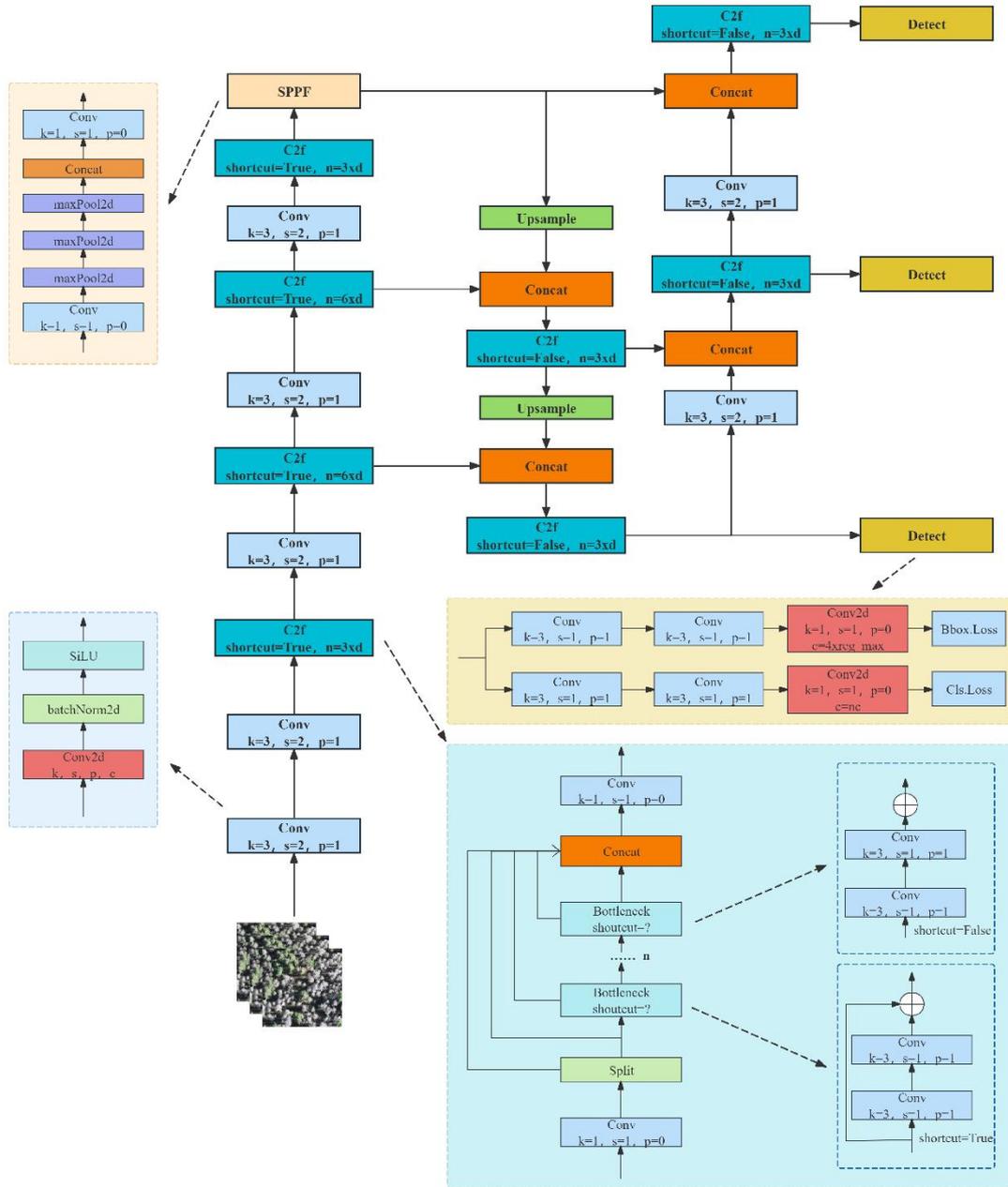

**Fig. 7.** YOLOv8 model architecture.

FID-Net builds upon YOLOv8 by introducing two custom-designed modules—FEM and AMFM—and incorporating the ECA attention mechanism. The overall architecture is illustrated in Fig. 8. Specifically, the lightweight FEM explicitly extracts disease-sensitive cues from a single RGB image and jointly models them with the original visual information. The network adopts a dual-branch backbone: the original RGB image is first processed by the FEM to generate a set of task-oriented feature images (TOFI) with three channels. Subsequently, both the RGB image and TOFI are fed into two parallel yet asymmetric feature extraction branches—the main branch preserves high-dimensional color and texture details, whereas the auxiliary branch focuses on spectral anomalies and structural degradation cues. Multi-scale features from both branches are aligned and adaptively fused using AMFM. Meanwhile, ECA modules are embedded at critical nodes of the network. Finally, the detection head outputs the localization coordinates and classification labels for infected and healthy trees.

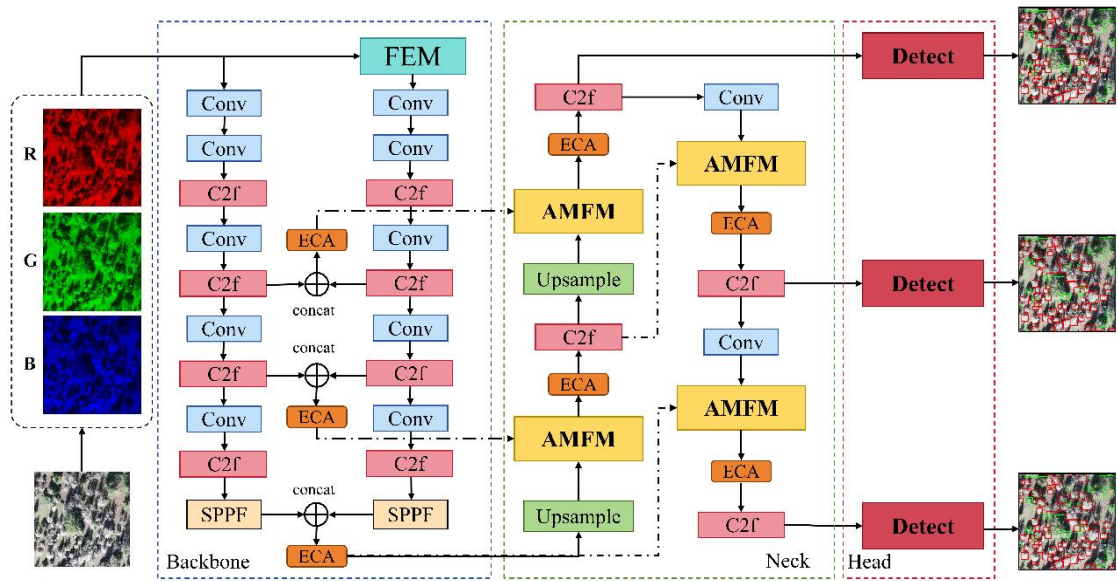

**Fig. 8.** FID-Net architecture.

(1) Feature Enhancement Module (FEM)

In forestry remote sensing scenarios, healthy and diseased vegetation exhibit significant differences in spectral response in the near-infrared band; however, low-cost consumer-grade UAVs typically only carry RGB cameras. To improve the detection accuracy of infected trees using only cameras, FEM was proposed to serve as a front-end preprocessing unit. Taking the RGB image as input, FEM applies lightweight image transformations to explicitly generate a set of task-oriented feature images (TOFI) that supplement the spectral anomaly cues and structural degradation information that RGB images fail to provide directly.

Specifically, FEM calculates three parallel channels to form a three-channel enhanced feature tensor, maintaining the same spatial resolution as the input:

Channel 0: Visible-band Difference Vegetation Index (VDVI)(Xu et al., 2020). This index simulates the operational logic of Normalized Difference Vegetation Index (NDVI) within the visible spectrum by leveraging the strong reflection of green light and the absorption characteristics of red and blue light in healthy vegetation. It highlights canopy color abnormalities caused by chlorophyll loss. The index is computed as:

$$VDVI = \frac{2G - R - B}{2G + R + B} \tag{1}$$

Channel 1: Laplacian texture map. Generated by computing the Laplacian operator of the grayscale image and applying absolute values, this channel encodes color-independent morphological information. It is crucial for delineating tree-crown contours and identifying structural degradation resulting from disease. The operator is defined as:

$$T(x,y) = \left|\nabla^2 I(x,y)\right| = \left|\frac{\partial^2 I}{\partial x^2} + \frac{\partial^2 I}{\partial y^2}\right| \tag{2}$$

Channel 2: Normalized Green–Blue Difference Index (NGBDI)(X. Wang et al., 2023). This index exploits the strong blue-band absorption of healthy vegetation. When a tree becomes infected, chlorophyll degradation reduces blue-light absorption while enhancing reflectance–NGBDI amplifies this spectral difference. The index is computed as follows:

$$NGBDI = \frac{(R_{Green} - R_{Blue})}{(R_{Green} + R_{Blue})} \quad (3)$$

FEM calculates the three channel indices, which are subsequently fed into the downstream dual-branch backbone network. The RGB branch preserves color information and fine-grained details, while the FEM branch emphasizes disease-related spectral–structural anomalies.

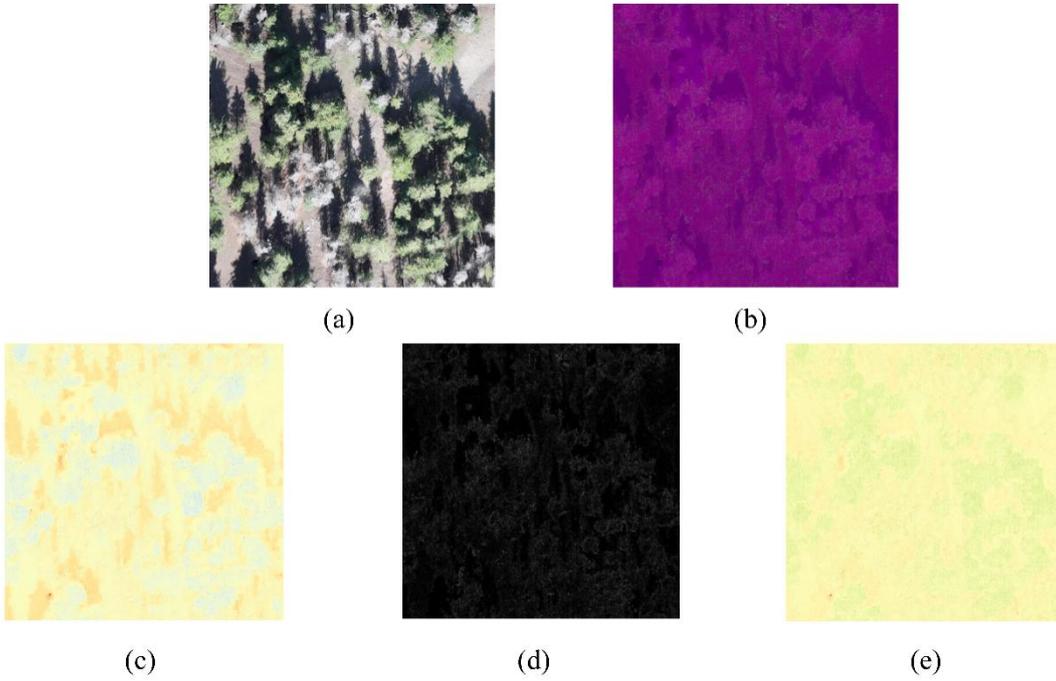

**Fig. 9.** (a) Original RGB image; (b) Task-oriented feature images; (c) VDVI output; (d) Texture feature map; (e) NGBDI output.

As **Fig. 9** shows, VDVI and NGBDI explicitly extract disease-sensitive cues, effectively distinguishing infected trees from healthy ones and differentiating background elements such as shadows or grasslands from true tree crowns. These indices significantly enhance the model's sensitivity to early-stage loss of greenness and maintain strong robustness under challenging illumination conditions such as strong sunlight, cloudy conditions, or shadows. In contrast, the Laplacian texture map focuses on local canopy sharpness and structural integrity, highlighting structural deterioration such as sparse foliage or branch dieback. This property is essential for accurately delineating crown boundaries in complex backgrounds and substantially improves the model's perception of morphological abnormalities. This design eliminates the need for additional sensors in traditional multimodal approaches and fully integrates feature enhancement into the front end of the model, enabling seamless end-to-end training and deployment.

(2) Adaptive Multi-scale Feature Fusion Module (AMFM)

In the dual-branch architecture, features extracted by the RGB and FEM branches exhibit substantial differences in semantic characteristics. The RGB branch contains natural color information and high-frequency texture details, whereas the FEM branch encodes disease-sensitive cues derived from physical priors. Direct channel concatenation can lead to semantic misalignment and inter-modal interference.

As shown in Fig. 10, AMFM first projects features from both branches at each scale into a unified semantic space via independent 1×1 convolutions, ensuring channel alignment. Subsequently, learnable scalar weights normalized by a softmax function are applied to perform weighted fusion of the two modalities. This mechanism enables the network to dynamically modulate the contribution of each modality according to target scale and scene context—for example, relying more on spectral anomaly cues from the FEM branch for small distant targets, and emphasizing fine-grained RGB details for nearby large crowns.

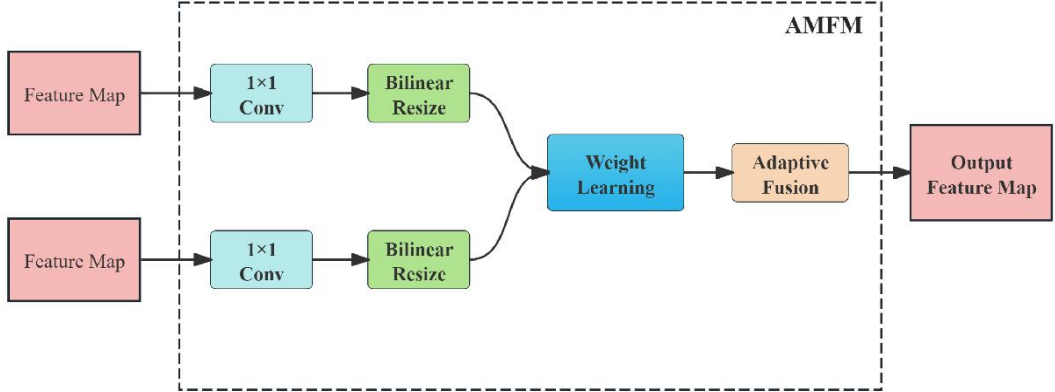

**Fig. 10.** AMFM module architecture.

(3) Efficient Channel Attention (ECA)

After dual-branch extraction and AMFM fusion, enhancing critical channel responses and suppressing noise is pivotal to improving detection performance. Traditional channel-attention mechanisms compress spatial information using global average pooling followed by fully connected layers, which increases computational cost and risks the loss of local details. In contrast, Efficient Channel Attention (ECA) replaces fully connected layers with a one-dimensional convolution that captures cross-channel interactions at very low computational cost(Wang et al., 2020).

Given that the discriminative cues of infected trees occur in specific FEM channels, this study applies ECA to recalibrate the fused feature maps. As shown in Fig. 11, the ECA module preserves spatial structure while strengthening disease-sensitive channels through an adaptive 1D convolution without dimensionality reduction.

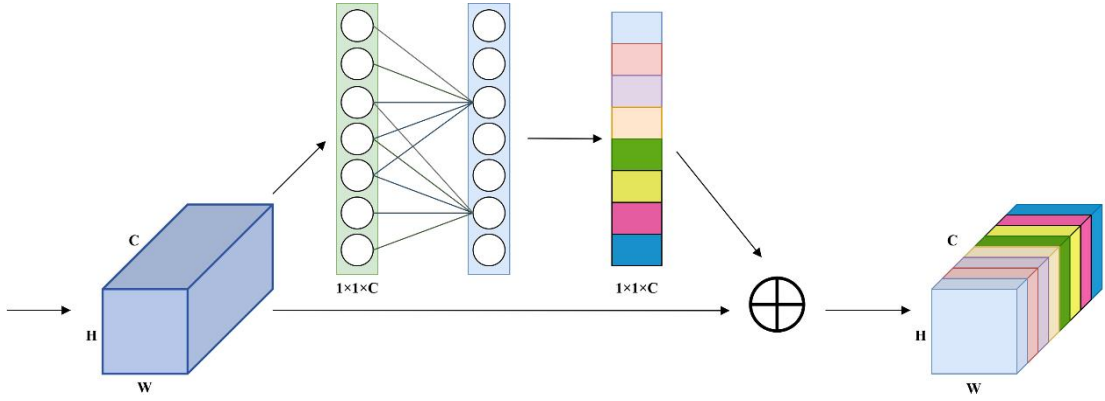

**Fig. 11.** ECA module architecture.

*2.3.3 Evaluation Metrics*

(1) Model evaluation metrics

To evaluate detection performance, all test images were manually annotated to identify infected and healthy trees. These annotations served as ground truth and were compared against the model's detection outputs to quantify performance.

The evaluation metrics included Precision (P), Recall (R), and mean Average Precision (mAP). Precision measures the correctness of positive predictions, i.e., the proportion of correctly detected infected and healthy trees among all predictions. Recall measures the proportion of actual infected and healthy trees that were correctly detected. Two mAP metrics were reported: mAP@0.5 (IoU threshold = 0.5) and mAP@0.5:0.95 (mean mAP across IoU thresholds from 0.5 to 0.95 with a step of 0.05).

$$Precision = \frac{TP}{TP+FP} \quad (4)$$

$$Recall = \frac{TP}{TP+FN} \quad (5)$$

(2) Ecological Analysis Methods

Based on the model's detection results of infected and healthy trees from the model, the experiment conducts a comprehensive analysis of pest and disease spread patterns from three perspectives: spatial distribution characteristics of infected trees, risk assessment of healthy trees, and identification of key protection areas.

First, the Kernel Density Estimation (KDE) method is employed to analyze the spatial distribution characteristics of infected trees. KDE is a classic non-parametric probability density function estimation technique, whose core principle involves assigning a smoothing kernel function to each data point; superimposing these kernel functions, yields the probability density distribution across the entire study area. This approach effectively captures the clustering patterns of spatial point distributions.

Assuming the set of location coordinates for infected trees is $P = \{(x_i, y_i)\}_{i=1}^{n}$, where $n$ represents the total number of infected trees, and $(x_i, y_i)$ denotes the center point of the $i$-th infected tree in the image coordinate system. The density estimate at any location $z = (x, y)$ is calculated as follows:

$$\hat{f}(z) = \frac{1}{nh^2} \sum_{i=1}^{m} K\left(\frac{\|z - z_i\|}{h}\right) \quad (6)$$

where:

$K(\cdot)$ is the kernel function. In this study, the Gaussian kernel is adopted:

$$K(u) = \frac{1}{2\pi} e^{-\frac{1}{2} u^T u} \quad (7)$$

$h$ is the bandwidth, which controls the smoothness of the kernel function.

$$h = n^{-\frac{1}{d+4}} \quad (8)$$

Here, $n$ denotes the number of infected trees, and $d$ represents the data dimension. In the experiment, $d = 2$, as the data includes two spatial dimensions: $x_i$ and $y_i$.

Finally, by computing the density estimates $\hat{f}(z)$ at grid points across the image plane and generating contour maps, an intuitive spatial distribution heatmap of infected trees is obtained.

Second, in pest-affected forest areas, healthy trees located in proximity to high-density infected tree clusters exhibit a correspondingly higher risk of infection. Thus, based on the KDE-derived spatial distribution heatmap of infected trees, this study further quantifies the risk index of each healthy tree. The specific steps are as follows:

1) Define the neighborhood scope. For each healthy tree, a pre-determined search radius $r$ is set, as determined by the typical dispersal distance of bark beetles.

2) Calculate the average neighborhood density. Within this neighborhood, the KDE-generated spatial distribution heatmap of infected trees is queried to compute the average density value, which represents the risk index for the target healthy tree.

For the $j$-th healthy tree, its risk score $R_j$ is defined as:

$$R_j = \frac{1}{|N_j|} \sum_{X_k \in N_j} \hat{f}(X_k) \qquad (9)$$

where:

$N_j$ denotes the set of all infected trees within the circular neighborhood centered on healthy tree $j$ with radius $r$.

$|N_j|$ denotes the number of infected trees in this neighborhood.

$\hat{f}(X_k)$ represents the disease density estimate at grid point $X_k$.

Through the above steps, a risk assessment map of healthy trees is generated. A higher risk index indicates greater pest and disease infection pressure in the neighborhood environment of the healthy tree, implying a elevated risk of subsequent infection.

Third, the experiment adopts the DBSCAN algorithm to identify regions with a high concentration of healthy trees within infected forests and delineates such regions with elliptical contours. The shape of the ellipse also reflects the density and extension direction of tree distribution within the region. These areas, exhibiting potential resistance to disease spread, should be prioritized for protective interventions and play a pivotal role in safeguarding the overall stability of the forest ecosystem.

## 3. Results

This section compares the performance of FID-Net against mainstream YOLO series models, analyzes the contributions of key modules such as FEM, AMFM, and ECA through ablation experiments, and finally conducts a situational analysis of pest and disease spread based on the detection results. This fully demonstrates the advantages of FID-Net in precise pest and disease detection and control.

### 3.1 Model Training Dynamics

During the experimental training phase, the following hyperparameters were adopted: an initial learning rate of 0.001 (cosine annealing scheduler), the AdamW optimizer with EMA (decay = 0.937), a batch size of 8, and a total of 200 training epochs. A comparative analysis of the training processes between FID-Net and YOLOv8n is presented in Fig. 12.

Both models exhibited a steady decrease in training and validation loss as the number of epochs increased, with losses stabilizing after approximately 150 epochs—indicating effective convergence for both. However, FID-Net exhibited lower loss values from the early stages of training, and its final convergence level outperformed that of YOLOv8n. This demonstrates that the proposed modules effectively not only accelerated model learning but also elevated the performance upper bound. Furthermore, the training and validation loss curves of FID-Net showed strong consistency, with no significant overfitting observed.

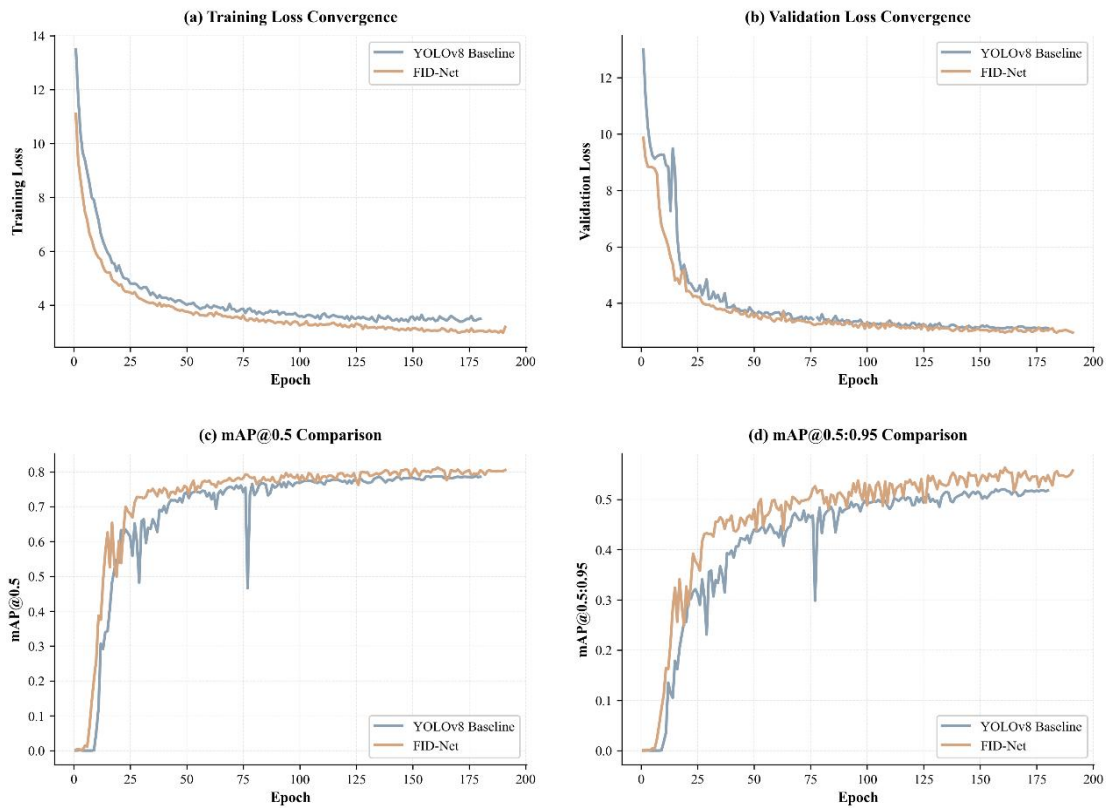

**Fig. 12.** Comparison of training processes between FID-Net and the baseline model YOLOv8n.

## 3.2 Comparison of Results from Different Models

To verify the comprehensive performance superiority of FID-Net, a systematic comparison was conducted with state-of-the-art YOLO-series object detection models under consistent experimental settings. Table 1 summarizes the detailed evaluation results. Evidently, FID-Net outperforms the comparison models across all key metrics. Specifically, its mAP@0.5 reached 82.29%, representing an improvement of 2.1 percentage points over the baseline model YOLOv8n. On the stricter mAP@0.5:0.95 metric, the performance gain is even more significant, reaching 64.30%, which is 4.66 percentage points higher than YOLOv8n. Notably, while newer generation models such as YOLOv10n, YOLOv11n, and YOLOv12n have shown outstanding performance in other domains, they failed to surpass the optimized YOLOv8n baseline in the task of detecting infected trees in complex forest areas. This underscores that scenario-specific structural optimizations offer greater practical significance than simply chasing the latest model iterations.

**Table 1.** Evaluation Results of Different Models

| Model | Precision (%) | Recall (%) | mAP@0.5 (%) | mAP@0.5:0.95 (%) |
|---|---|---|---|---|
| YOLOv5n | 83.58 | 71.81 | 79.69 | 58.70 |
| YOLOv10n | 76.62 | 68.73 | 74.19 | 52.68 |
| YOLOv11n | 82.03 | 72.37 | 79.41 | 58.89 |
| YOLOv12n | 79.88 | 70.98 | 77.65 | 55.38 |
| YOLOv8n | 83.87 | 72.29 | 80.19 | 59.64 |
| **FID-Net(ours)** | **86.10** | **75.44** | **82.29** | **64.30** |

In terms of precision, FID-Net achieved a value of 86.10%, indicating a low false detection rate. Simultaneously, the model's recall rate reached 75.44%, demonstrating the model's strong coverage capability for real infected and healthy tree samples. This result reflects that FID-Net maintains excellent detection robustness under highly cluttered forest backgrounds—characterized by terrain undulations, lighting variations, and mixed vegetation cover—effectively striking a balance between detection precision and recall.

Fig. 13 presents the individual tree-level detection results for Sample Plot 1 and Sample Plot 2. Red bounding boxes indicate infected trees, while green bounding boxes indicate healthy trees. These two sample plots not only cover a wide spatial area but also contain a large number of densely distributed dead or infected trees: Sample Plot 1 contains a total of 5,666 trees, of which 2,818 are infected; Sample Plot 2 contains 6,993 trees, with 3,602 infected. Under such conditions, FID-Net still accurately distinguished between infected and healthy trees, achieving stable and precise localization.

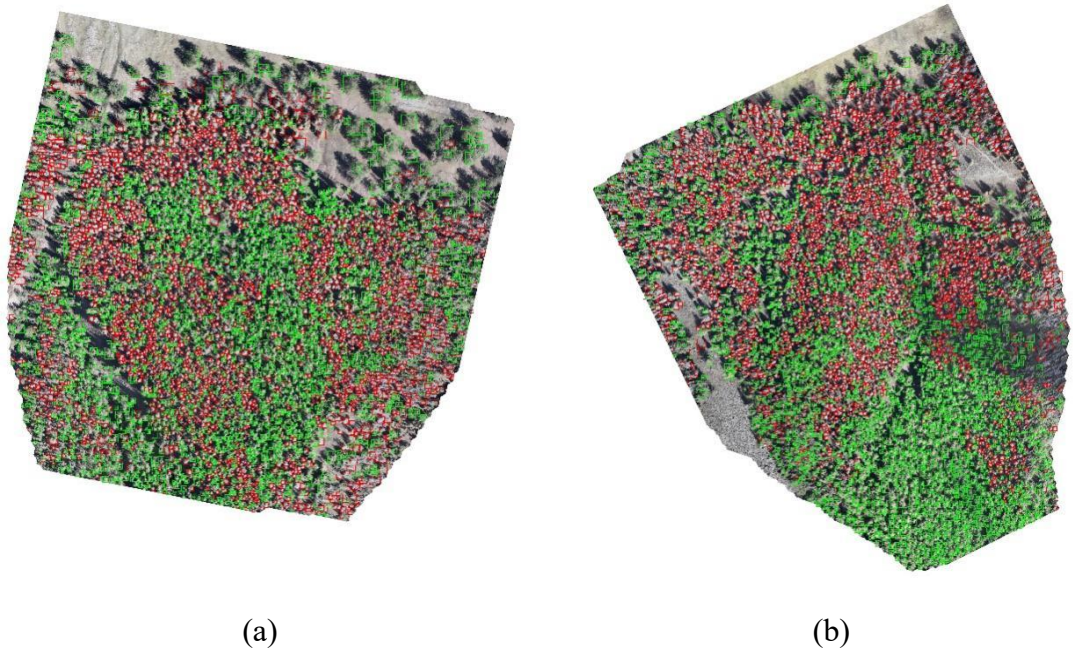

(a)             (b)

**Fig. 13.** Sample plot detection results. (a) Sample Plot 1; (b) Sample Plot 2.

## 3.3 Ablation Study

To systematically verify the individual effectiveness and synergistic effects of the proposed FEM, AMFM and ECA modules, a series of comprehensive ablation experiments were carried out on the YOLOv8n baseline model. All experiments were performed under the identical training configuration, with results presented in Table 2.

**Table 2.** Ablation Study Results of the FID-Net.

| Model | Precision (%) | Recall (%) | mAP@0.5 (%) | mAP@0.5:0.95 (%) |
|---|---|---|---|---|
| YOLOv8n | 83.87 | 72.29 | 80.19 | 59.64 |
| +FEM | 85.07 | 74.29 | 81.20 | 62.39 |
| FEM+ AMFM | 85.01 | 74.76 | 81.59 | 62.69 |
| FEM+ AMFM + CA | 84.43 | 74.71 | 81.49 | 62.61 |
| FEM+ AMFM + CBAM | 84.84 | 74.54 | 81.42 | 62.53 |
| FEM+ AMFM + SimAM | 85.31 | 75.40 | 82.12 | 64.25 |
| **FEM+ AMFM + ECA（FID-Net）** | **86.10** | **75.44** | **82.29** | **64.30** |

Obviously, FEM yeilds a substantial model performance: mAP@0.5:0.95 rises from the baseline of 59.64% to 62.39%, representing a gain of 2.75 percentage. This validates that lightweight transformations, such as VDVI, Laplacian texture maps, and NGBDI, can effectively compensate the limitations of the RGB modality in disease-sensitive information.

AMFM was employed to replace straightforward feature concatenation. Experimental results show that the model's recall rate increases from 74.29% to 74.76%, while mAP@0.5 and mAP@0.5:0.95 also rises to 81.59% and 62.69%, respectively. Therefore, the effectiveness of AMFM was validated in cross-modal semantic alignment and dynamic fusion.

ECA was compared with other mainstream attention mechanisms including Channel Attention (CA), Convolutional Block Attention Module (CBAM), and the parameter-free attention module SimAM. As shown in Table 4, under the identical architecture, ECA performed best across all evaluation metrics. Although CA and CBAM provided marginal improvements, their intricate structures fail to translate into performance benefits. While SimAM performed reasonably well in terms of precision (85.31%) and mAP@0.5 (82.12%), its recall rate was only 75.40%, which is lower than other schemes. These results further highlight the comprehensive advantage of ECA in maintaining high precision while also ensuring high recall capability.

The final FID-Net model achieved optimal performance across all metrics: precision increased to 86.10%, recall reached 75.44%, and mAP@0.5 and mAP@0.5:0.95 reached 82.29% and 64.30%, respectively.

## 3.4 Detection Results Across Different Scenarios

To evaluate the model's adaptability in diverse environments, Figure 14 presents the detection results of FID-Net in different typical scenarios. In the figure, red boxes indicate detected infected trees, green boxes indicate detected healthy trees, and yellow areas highlight the comparison regions between the FID-Net model and the baseline model.

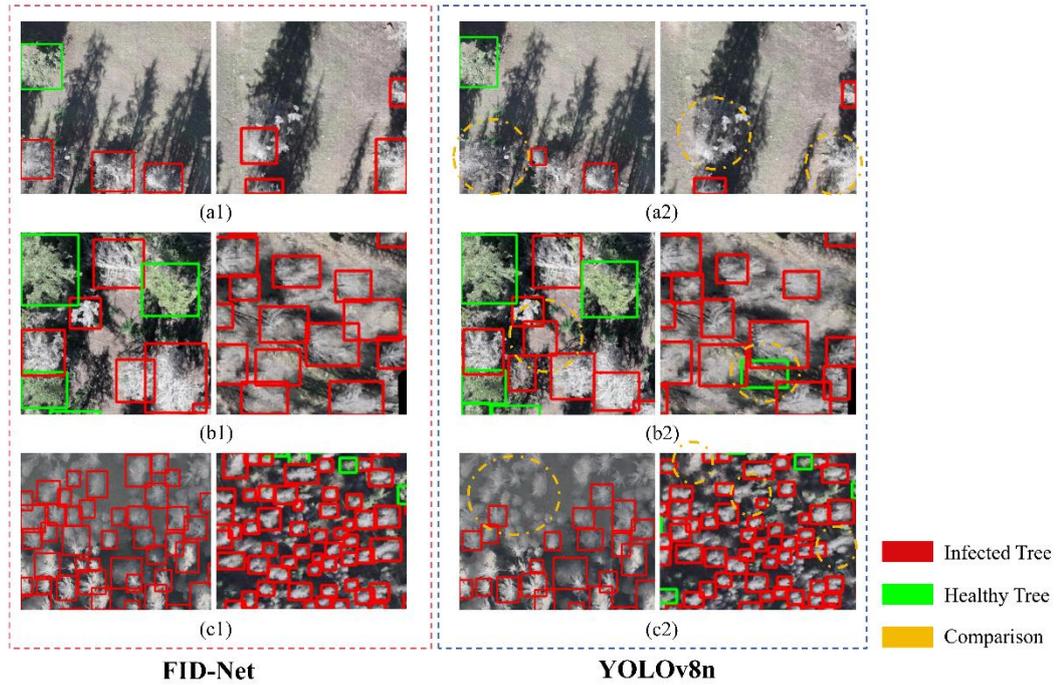

**Fig. 14.** Detection results under different scenarios.

Fig. 14(a1) and Fig. 14(a2) demonstrate the model's detection performance for close-range large targets and targets under shadows. As observed from the yellow circles in the two subfigures of Fig. 14(a2), the model can accurately identify and localize infected regions even under close-range and shadowed conditions.

As shown in Fig. 14(b1) and Fig. 14(b2), in scenarios with excessively strong illumination or ground-level green vegetation, the baseline model is often misled, mistakenly identifying illuminated ground as infected trees or green grassland as healthy trees.

Regarding the detection of small, distant aerial targets, Fig. 14(c1) and Fig. 14(c2) further present the model's results for small infected regions under long-distance, high-altitude conditions. Particularly in large-scale infected or shadowed regions, the detection task becomes more complex due to background variations and increased distance. Nevertheless, the FID-Net model still performed well, effectively identifying and annotating infected areas, demonstrating its strong potential for practical application.

## 3.5 Situational Analysis of Tree Infection Status

The detected infected and healthy trees were spatially visualized in geographical space, with the results presented in Fig. 15. Infected trees exhibited clustered distribution characteristics across both sample plots, forming multiple locally dense regions.

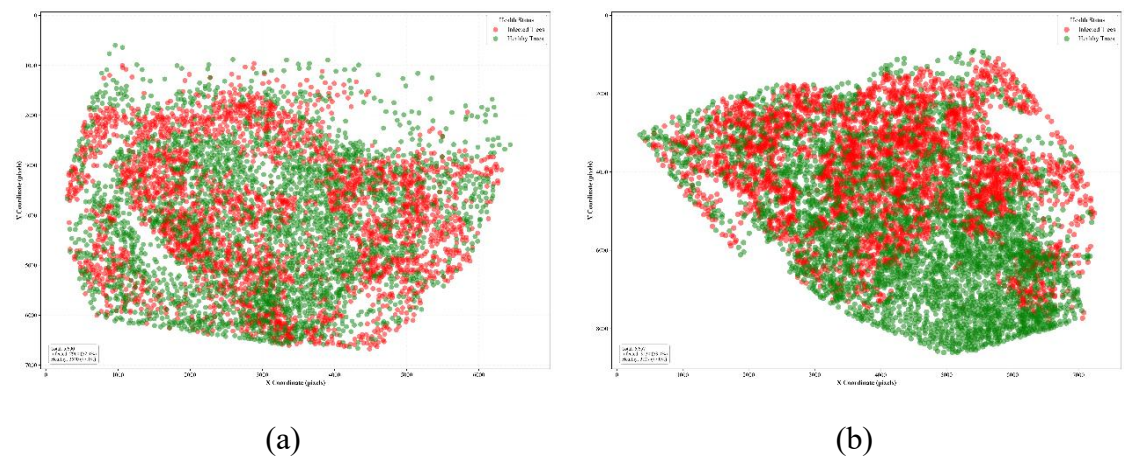

(a)                  (b)

**Fig. 15.** Spatial distribution characteristics of infected and healthy trees. (a) Sample Plot 1; (b) Sample Plot 2.

By using KDE, a continuous spatial density field of the infected trees was constructed, with results presented in Fig. 16. The density maps reveal that the infected trees form multiple high-intensity core regions across both sample plots, which exhibit clear spatial boundaries and gradient structures. In these figures, darker-colored regions represent higher densities of infected trees, indicating potential disease outbreak or transmission core zones.

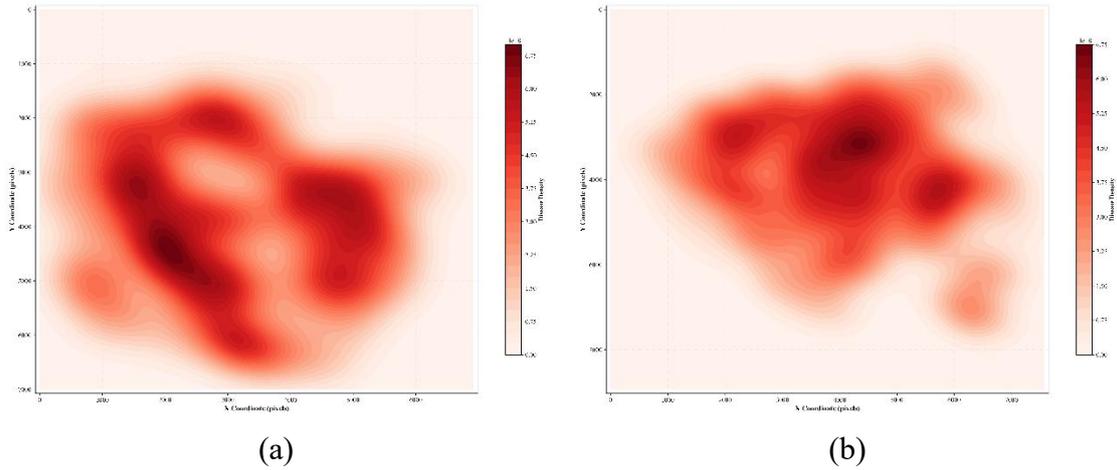

**Fig. 16.** Heatmaps of the spatial distribution of infected trees. (a) Sample Plot 1; (b) Sample Plot 2.

Fig. 17 shows the distribution of healthy trees within the spatial risk field. All healthy trees are represented by blue scatter points, with color intensity determined by their risk index: darker color indicates higher risk, lighter color indicates lower risk. The red gradient heatmap in the background represents the density field of infected trees.

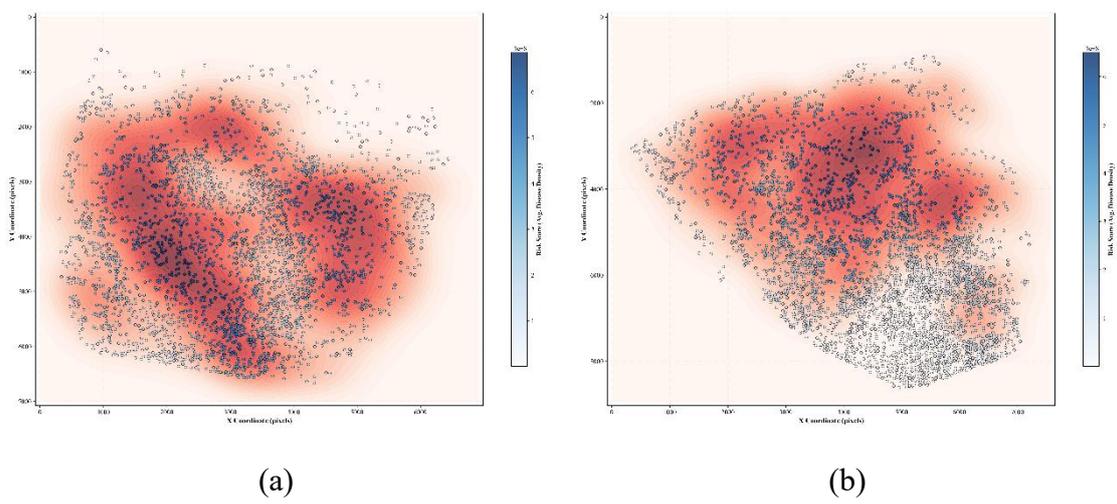

**Fig. 17.** Risk assessment maps for healthy trees. (a) Sample Plot 1; (b) Sample Plot 2.

The DBSCAN algorithm was applied to cluster healthy trees, identifying three high-density regions of healthy trees. As shown in Fig. 18, an ellipse was fitted to each cluster, with the three key protection areas labeled as PA1, PA2, and PA3.

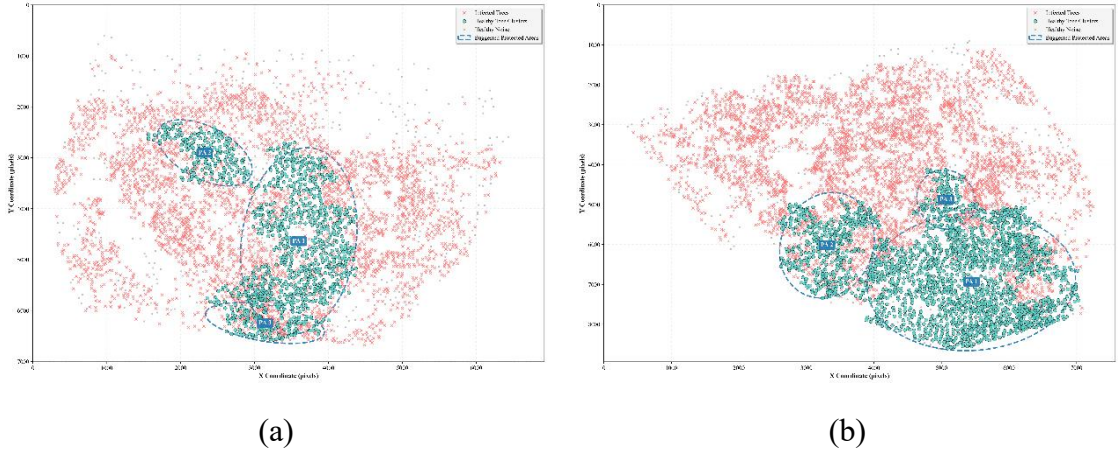

(a)                          (b)

**Fig. 18.** Maps of key protection areas in the sample plots. (a) Sample Plot 1; (b) Sample Plot 2.

## 4. Discussion

This study successfully constructed and systematically validated the FID-Net model based on consumer-grade UAV visible-light imagery, achieving high-precision simultaneous identification of infected and healthy trees in the forests of the East Tianshan Mountains in Xinjiang. Experimental results demonstrate that the performance metrics achieved by this model are significantly superior to those of mainstream YOLO series models. The organic synergy among the FEM, the AMFM, and the ECA mechanism within the FID-Net model effectively addresses for the inherent limitations of the single RGB modality in capturing disease-sensitive information, thereby enhancing the feature discriminability between infected and healthy trees.

Furthermore, compared with existing forest disease detection research, FID-Net achieves three key breakthroughs: First, the model specifically optimizes for core technical challenges in high-density infection scenarios, such as target crowding, significant scale heterogeneity, and class imbalance, significantly improving detection applicability in complex forest scenes. Second, it innovatively realizes the simultaneous and accurate identification of both infected and healthy trees. This not only locates disease occurrence areas but also comprehensively captures the spatial distribution of healthy trees within the stand, providing crucial data support for the systematic assessment of the overall stand health status. Third, it achieves high-performance detection relying solely on low-cost, consumer-grade UAV visible-light imagery, effectively circumventing the application limitations of traditional multispectral/hyperspectral equipment, such as high costs, complex operation, and difficult data processing. These characteristics make the FID-Net model a more feasible and widely applicable technical solution for grassroots forest management departments.

In addition, by integrating the high-precision detection results with ecological analysis, risk assessment, and prediction, this study directly converts the outcomes into ecological decision-making support. It establishes a closed-loop transition from technological development to management implementation at the practical application level, significantly enhancing the practical value of the research.

This study analyzed the spatial distribution pattern of bark beetle infestation in the East Tianshan forest area. Results indicate that infected trees exhibit a distinct "clustered" aggregation pattern—consistent with the biological traits and dispersal mechanisms of bark beetles. Existing research confirms that after successfully infesting a host tree, bark beetles release specific aggregation pheromones, attracting conspecifics to launch mass attacks on nearby healthy trees, thereby forming spatially cohesive disease diffusion cores(Blomquist et al., 2010; Ramakrishnan et al., 2025).

This study visually presents this aggregation characteristic through the KDE method, not only providing empirical support for the ecological theory of bark beetle disease spread but also clearly delineating the core areas and extent of disease diffusion, offering a spatial basis for precise control measures.

Notably, as typical wood-boring pests, bark beetles tend to preferentially infest small-diameter trees with vigorous growth, thinner bark, and tender xylem tissues as their initial hosts. Such trees possess relatively weaker defense capabilities, making them more susceptible to bark beetles penetrating the cortical barrier. In contrast, large-diameter trees are typically older, with thicker cortical tissues, higher lignification, and more robust defense systems, granting them stronger resistance to bark beetle infestation(Jaime et al., 2019).

In the experiment, tree diameter classes in the sample plots were further categorized. Since UAVs capture images of the same plot with consistent camera parameters and shooting distances, the crown size in the imagery is positively correlated with the actual crown size of trees. Taking Sample Plot 1 and Sample Plot 2 as examples, the crown sizes of all detected trees were calculated and divided into three equal intervals: large, medium, and small. Fig. 19 presents the histograms of tree crown area distribution for the two sample plots, where the red dashed line indicates the boundary between small and medium diameter classes, and the blue dashed line indicates the boundary between medium and large diameter classes.

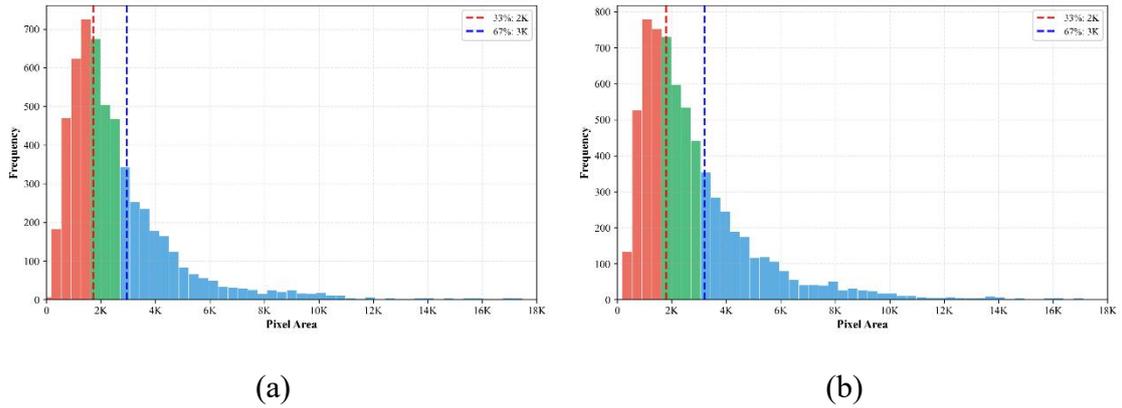

(a) (b)

**Fig. 19.** (a) Histogram of bounding box pixel area distribution for Sample Plot 1; (b) Histogram of bounding box pixel area distribution for Sample Plot 2.

Statistical analysis of healthy and infected trees across different diameter classes in both sample plots is shown in Fig. 20. The red portion represents the number of infected trees, the green portion represents the number of healthy trees, and the number above each bar indicates the total tree count for that diameter class. Clearly, both plots show a consistent trend: small-diameter trees have the highest infection proportion, followed by medium-diameter trees, while large-diameter trees have the lowest proportion of infection. This experimental result is consistent with the ecological mechanisms underlying bark beetle dispersal.

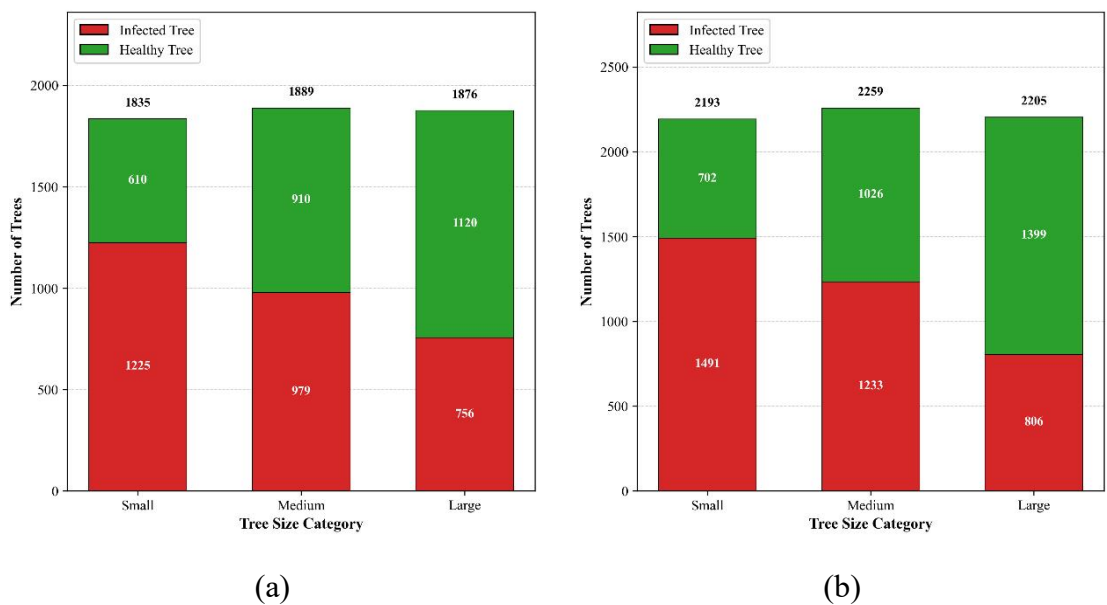

(a) (b)

**Fig. 20.** Statistics of disease occurrence in trees of different size classes. (a) Sample Plot 1; (b)

Sample Plot 2

This study also transcends the limitation of traditional methods that solely focus on individual tree disease statuses, highlighting the spatial continuity and neighborhood effects of pest infestations. By analyzing the spatial topological relationship between each healthy tree and neighboring infected trees, we established a model for predicting potential infestation directions and high-risk areas, achieving a shift from static monitoring to dynamic prediction. Prediction results show that the infection risk for healthy trees exhibits a clear spatial gradient structure, with risk levels showing a significant positive correlation with disease density. Consequently, the closer a healthy tree is to the disease core area and the higher the density of surrounding infected trees, the greater its risk of infection, warranting more intensive control attention.

While this study has achieved some results, further improvements of the model research and ecological analysis are required. During actual data collection, abnormal imaging conditions may occur, such as strong light, image blurring, and distortion. In these scenarios, the distinguishability of tree features is reduced, potentially reducing the precision of model feature extraction. Therefore, training with multi-temporal and multi-regional datasets is needed, which can further enhance the model's accuracy and robustness.

## 5. Conclusion

Forest pest and disease monitoring represents an extremely important yet systematically challenging task, constrained by multiple complex factors. This paper proposes the FID-Net model for monitoring tree mortality caused by bark beetle infestations in eastern Tianshan forest areas of Xinjiang, conducting precise detection of infected and healthy trees in forests based on UAV images, coupled with comprehensive pest infestation situation analysis.

Through the innovative FEM, AMFM Modules, and the introduction of the ECA mechanism, this study achieves high-precision detection of infected and healthy trees using only UAV visible-light imagery. Simultaneously, the detection results provided high-quality data support for subsequent pest situation analysis, realizing an integrated closed-loop from identification to assessment.

However, the occurrence of forest pest and disease is shaped by the interplay of multiple complex factors, including tree species types, topography, climate conditions, and human interference. Current research still has certain limitations. For example, data were collected during a single summer time period; model robustness under extreme lighting and image blur conditions requires further enhancement; and generalization ability across different regions and tree species requires further validation. Therefore, future research could expand the application scope of this technology to diverse ecological zones, various disease types, and multi-temporal scenarios, thereby continuously enhancing the model's environmental adaptability and robustness. In subsequent work, we will strengthen multi-source data research on forest pest monitoring and continue exploring precise protection and sustainable management of forest ecosystems.

## CRediT authorship contribution statement

Yan Zhang: Writing – original draft, Data curation, Visualization, Software, Methodology, Conceptualization. Baoxin Li: Writing – review & editing, Data curation, Visualization. Han Sun: Writing – review & editing, Funding acquisition, Investigation. Pei Wang: Writing – review & editing, Funding acquisition, Conceptualization, Supervision. Yuhang Gao : Investigation, Data curation. Mingtai Zhang: Investigation, Data curation.

## Declaration of competing interest

The authors declare that they have no known competing financial interests or personal relationships that could have appeared to influence the work reported in this paper.

## Acknowledgements

This research was funded by the Third Xinjiang Scientific Expedition Program (Grant No.2022xjkk1205) and supported by the Beijing Natural Science Foundation (No. 6232031) .

## Data availability

Data will be made available on request.

## References

Blomquist, G.J., Figueroa-Teran, R., Aw, M., Song, M., Gorzalski, A., Abbott, N.L., Chang, E., Tittiger, C., 2010. Pheromone production in bark beetles. Insect Biochemistry and Molecular Biology 40, 699–712. https://doi.org/10.1016/j.ibmb.2010.07.013

Brovkina, O., Cienciala, E., Surový, P., Janata, P., 2018. Unmanned aerial vehicles (UAV) for assessment of qualitative classification of Norway spruce in temperate forest stands. Geo-spatial Information Science 21, 12–20. https://doi.org/10.1080/10095020.2017.1416994

Calvin, K., Dasgupta, D., Krinner, G., Mukherji, A., Thorne, P.W., Trisos, C., Romero, J., Aldunce, P., Barrett, K., Blanco, G., Cheung, W.W.L., Connors, S., Denton, F., Diongue-Niang, A., Dodman, D., Garschagen, M., Geden, O., Hayward, B., Jones, C., Jotzo, F., Krug, T., Lasco, R., Lee, Y.-Y., Masson-Delmotte, V., Meinshausen, M., Mintenbeck, K., Mokssit, A., Otto, F.E.L., Pathak, M., Pirani, A., Poloczanska, E., Pörtner, H.-O., Revi, A., Roberts, D.C., Roy, J., Ruane, A.C., Skea, J., Shukla, P.R., Slade, R., Slangen, A., Sokona, Y., Sörensson, A.A., Tignor, M., Van Vuuren, D., Wei, Y.-M., Winkler, H., Zhai, P., Zommers, Z., Hourcade, J.-C., Johnson, F.X., Pachauri, S., Simpson, N.P., Singh, C., Thomas, A., Totin, E., Arias, P., Bustamante, M., Elgizouli, I., Flato, G., Howden, M., Méndez-Vallejo, C., Pereira, J.J., Pichs-Madruga, R., Rose, S.K., Saheb, Y., Sánchez Rodríguez, R., Ürge-Vorsatz, D., Xiao, C., Yassaa, N., Alegría, A., Armour, K., Bednar-Friedl, B., Blok, K., Cissé, G., Dentener, F., Eriksen, S., Fischer, E., Garner, G., Guivarch, C., Haasnoot, M., Hansen, G., Hauser, M.,

Hawkins, E., Hermans, T., Kopp, R., Leprince-Ringuet, N., Lewis, J., Ley, D., Ludden, C., Niamir, L., Nicholls, Z., Some, S., Szopa, S., Trewin, B., Van Der Wijst, K.-I., Winter, G., Witting, M., Birt, A., Ha, M., Romero, J., Kim, J., Haites, E.F., Jung, Y., Stavins, R., Birt, A., Ha, M., Orendain, D.J.A., Ignon, L., Park, S., Park, Y., Reisinger, A., Cammaramo, D., Fischlin, A., Fuglestvedt, J.S., Hansen, G., Ludden, C., Masson-Delmotte, V., Matthews, J.B.R., Mintenbeck, K., Pirani, A., Poloczanska, E., Leprince-Ringuet, N., Péan, C., 2023. IPCC, 2023: Climate Change 2023: Synthesis Report. Contribution of Working Groups I, II and III to the Sixth Assessment Report of the Intergovernmental Panel on Climate Change [Core Writing Team, H. Lee and J. Romero (eds.)]. IPCC, Geneva, Switzerland. Intergovernmental Panel on Climate Change (IPCC). https://doi.org/10.59327/IPCC/AR6-9789291691647

Canadell, J.G., Raupach, M.R., 2008. Managing Forests for Climate Change Mitigation. Science 320, 1456–1457. https://doi.org/10.1126/science.1155458

Chen, Z., Lin, H., Bai, D., Qian, J., Zhou, H., Gao, Y., 2025. PWDViTNet: A lightweight early pine wilt disease detection model based on the fusion of ViT and CNN. Computers and Electronics in Agriculture 230, 109910. https://doi.org/10.1016/j.compag.2025.109910

Dash, J.P., Watt, M.S., Pearse, G.D., Heaphy, M., Dungey, H.S., 2017. Assessing very high resolution UAV imagery for monitoring forest health during a simulated disease outbreak. ISPRS Journal of Photogrammetry and Remote Sensing 131, 1–14. https://doi.org/10.1016/j.isrsjprs.2017.07.007
44

Deng, X., Tong, Z., Lan, Y., Huang, Z., 2020. Detection and Location of Dead Trees with Pine Wilt Disease Based on Deep Learning and UAV Remote Sensing. AgriEngineering 2, 294–307. https://doi.org/10.3390/agriengineering2020019

Dong, X., Zhang, L., Xu, C., Miao, Q., Yao, J., Liu, F., Liu, H., Lu, Y.-B., Kang, R., Song, B., 2024. Detection of pine wilt disease infected pine trees using YOLOv5 optimized by attention mechanisms and loss functions. Ecological Indicators 168, 112764. https://doi.org/10.1016/j.ecolind.2024.112764

Harris, N.L., Gibbs, D.A., Baccini, A., Birdsey, R.A., de Bruin, S., Farina, M., Fatoyinbo, L., Hansen, M.C., Herold, M., Houghton, R.A., Potapov, P.V., Suarez, D.R., Roman-Cuesta, R.M., Saatchi, S.S., Slay, C.M., Turubanova, S.A., Tyukavina, A., 2021. Global maps of twenty-first century forest carbon fluxes. Nature Climate Change 11, 234–240. https://doi.org/10.1038/s41558-020-00976-6

Home | Global Forest Resources Assessments | Food and Agriculture Organization of the United Nations [WWW Document], 2025. . GlobalFRA. URL https://www.fao.org/forest-resources-assessment/en (accessed 12.6.25).

Hu, G., Yao, P., Wan, M., Bao, W., Zeng, W., 2022a. Detection and classification of diseased pine trees with different levels of severity from UAV remote sensing images. Ecological Informatics 72, 101844. https://doi.org/10.1016/j.ecoinf.2022.101844

Hu, G., Yin, C., Wan, M., Zhang, Y., Fang, Y., 2020. Recognition of diseased Pinus trees in UAV images using deep learning and AdaBoost classifier. Biosystems

Engineering 194, 138–151. https://doi.org/10.1016/j.biosystemseng.2020.03.021

Hu, G., Zhu, Y., Wan, M., Bao, W., Zhang, Y., Liang, D., Yin, C., 2022b. Detection of diseased pine trees in unmanned aerial vehicle images by using deep convolutional neural networks. Geocarto International 37, 3520–3539. https://doi.org/10.1080/10106049.2020.1864025

Huo, L., Lindberg, E., Bohlin, J., Persson, H.J., 2023. Assessing the detectability of European spruce bark beetle green attack in multispectral drone images with high spatial- and temporal resolutions. Remote Sensing of Environment 287, 113484. https://doi.org/10.1016/j.rse.2023.113484

Iordache, M.-D., Mantas, V., Baltazar, E., Pauly, K., Lewyckyj, N., 2020. A Machine Learning Approach to Detecting Pine Wilt Disease Using Airborne Spectral Imagery. Remote Sensing 12, 2280. https://doi.org/10.3390/rs12142280

Jaime, L., Batllori, E., Margalef-Marrase, J., Pérez Navarro, M.Á., Lloret, F., 2019. Scots pine (*Pinus sylvestris* L.) mortality is explained by the climatic suitability of both host tree and bark beetle populations. Forest Ecology and Management 448, 119–129. https://doi.org/10.1016/j.foreco.2019.05.070

Junttila, S., Näsi, R., Koivumäki, N., Imangholiloo, M., Saarinen, N., Raisio, J., Holopainen, M., Hyyppä, H., Hyyppä, J., Lyytikäinen-Saarenmaa, P., Vastaranta, M., Honkavaara, E., 2022. Multispectral Imagery Provides Benefits for Mapping Spruce Tree Decline Due to Bark Beetle Infestation When Acquired Late in the Season. Remote Sensing 14, 909. https://doi.org/10.3390/rs14040909

Lausch, A., Borg, E., Bumberger, J., Dietrich, P., Heurich, M., Huth, A., Jung, A.,

Klenke, R., Knapp, S., Mollenhauer, H., Paasche, H., Paulheim, H., Pause, M., Schweitzer, C., Schmulius, C., Settele, J., Skidmore, A.K., Wegmann, M., Zacharias, S., Kirsten, T., Schaepman, M.E., 2018. Understanding Forest Health with Remote Sensing, Part III: Requirements for a Scalable Multi-Source Forest Health Monitoring Network Based on Data Science Approaches. Remote Sensing 10, 1120. https://doi.org/10.3390/rs10071120

Li, C., Li, K., Ji, Y., Xu, Z., Gu, J., Jing, W., 2024. A spatio-temporal multi-scale fusion algorithm for pine wood nematode disease tree detection. J. For. Res. 35, 109. https://doi.org/10.1007/s11676-024-01754-2

Li, F., Liu, Z., Shen, W., Wang, Yan, Wang, Yunlu, Ge, C., Sun, F., Lan, P., 2021. A Remote Sensing and Airborne Edge-Computing Based Detection System for Pine Wilt Disease. IEEE Access 9, 66346–66360. https://doi.org/10.1109/ACCESS.2021.3073929

Liu, W., Guo, Z., Lu, F., Wang, X., Zhang, M., Liu, B., Wei, Y., Cui, L., Luo, Y., Zhang, L., Ouyang, Z., Yuan, Y., 2020. The influence of disturbance and conservation management on the greenhouse gas budgets of China's forests. Journal of Cleaner Production 261, 121000. https://doi.org/10.1016/j.jclepro.2020.121000

Netherer, S., Lehmanski, L., Bachlehner, A., Rosner, S., Savi, T., Schmidt, A., Huang, J., Paiva, M.R., Mateus, E., Hartmann, H., Gershenzon, J., 2024. Drought increases Norway spruce susceptibility to the Eurasian spruce bark beetle and its associated fungi. New Phytologist 242, 1000–1017.

https://doi.org/10.1111/nph.19635

Oblinger, B.W., Bright, B.C., Hanavan, R.P., Simpson, M., Hudak, A.T., Cook, B.D., Corp, L.A., 2022. Identifying conifer mortality induced by Armillaria root disease using airborne lidar and orthoimagery in south central Oregon. Forest Ecology and Management 511, 120126. https://doi.org/10.1016/j.foreco.2022.120126

Oide, A.H., Nagasaka, Y., Tanaka, K., 2022. Performance of machine learning algorithms for detecting pine wilt disease infection using visible color imagery by UAV remote sensing. Remote Sensing Applications: Society and Environment 28, 100869. https://doi.org/10.1016/j.rsase.2022.100869

Pan, Y., Birdsey, R.A., Fang, J., Houghton, R., Kauppi, P.E., Kurz, W.A., Phillips, O.L., Shvidenko, A., Lewis, S.L., Canadell, J.G., Ciais, P., Jackson, R.B., Pacala, S.W., McGuire, A.D., Piao, S., Rautiainen, A., Sitch, S., Hayes, D., 2011. A Large and Persistent Carbon Sink in the World's Forests. Science 333, 988–993. https://doi.org/10.1126/science.1201609

Qin, B., Sun, F., Shen, W., Dong, B., Ma, S., Huo, X., Lan, P., 2023. Deep Learning-Based Pine Nematode Trees' Identification Using Multispectral and Visible UAV Imagery. Drones 7, 183. https://doi.org/10.3390/drones7030183

Ramakrishnan, R., Shewale, M.K., Strádal, J., Frühbrodt, T., Doležal, P., Um-e-Hani, Andersson, M.N., Gershenzon, J., Jirošová, A., 2025. Aggregation Pheromones in the Bark Beetle Genus Ips: Advances in Biosynthesis, Sensory Detection, and Forest Management Applications. Curr. For. Rep. 11, 21.

https://doi.org/10.1007/s40725-025-00253-9

Sapkota, R., Ahmed, D., Karkee, M., 2024. Comparing YOLOv8 and Mask R-CNN for instance segmentation in complex orchard environments. Artificial Intelligence in Agriculture 13, 84–99. https://doi.org/10.1016/j.aiia.2024.07.001

Sharma, A., Kumar, V., Longchamps, L., 2024. Comparative performance of YOLOv8, YOLOv9, YOLOv10, YOLOv11 and Faster R-CNN models for detection of multiple weed species. Smart Agricultural Technology 9, 100648. https://doi.org/10.1016/j.atech.2024.100648

Singh, V.V., Naseer, A., Mogilicherla, K., Trubin, A., Zabihi, K., Roy, A., Jakuš, R., Erbilgin, N., 2024. Understanding bark beetle outbreaks: exploring the impact of changing temperature regimes, droughts, forest structure, and prospects for future forest pest management. Rev Environ Sci Biotechnol 23, 257–290. https://doi.org/10.1007/s11157-024-09692-5

Solimani, F., Cardellicchio, A., Dimauro, G., Petrozza, A., Summerer, S., Cellini, F., Renò, V., 2024. Optimizing tomato plant phenotyping detection: Boosting YOLOv8 architecture to tackle data complexity. Computers and Electronics in Agriculture 218, 108728. https://doi.org/10.1016/j.compag.2024.108728

The state of the world's forests 2024: forest-sector innovations towards a more sustainable future, 2024.

Wang, Q., Wu, B., Zhu, P., Li, P., Zuo, W., Hu, Q., 2020. ECA-Net: Efficient Channel Attention for Deep Convolutional Neural Networks, in: 2020 IEEE/CVF Conference on Computer Vision and Pattern Recognition (CVPR). Presented at the 2020 IEEE/CVF Conference on Computer Vision and Pattern Recognition (CVPR), pp. 11531–11539. https://doi.org/10.1109/CVPR42600.2020.01155

Wang, S., Cao, X., Wu, M., Yi, C., Zhang, Z., Fei, H., Zheng, H., Jiang, H., Jiang, Y., Zhao, Xianfeng, Zhao, Xiaojing, Yang, P., 2023. Detection of Pine Wilt Disease Using Drone Remote Sensing Imagery and Improved YOLOv8 Algorithm: A Case Study in Weihai, China. Forests 14, 2052. https://doi.org/10.3390/f14102052

Wang, X., Yan, S., Wang, W., Yin, L., Li, M., Yu, Z., Chang, S., Hou, F., 2023. Monitoring leaf area index of the sown mixture pasture through UAV multispectral image and texture characteristics. Computers and Electronics in Agriculture 214, 108333. https://doi.org/10.1016/j.compag.2023.108333

Wu, D., Yu, L., Yu, R., Zhou, Q., Li, J., Zhang, X., Ren, L., Luo, Y., 2023. Detection of the Monitoring Window for Pine Wilt Disease Using Multi-Temporal UAV-Based Multispectral Imagery and Machine Learning Algorithms. Remote Sensing 15, 444. https://doi.org/10.3390/rs15020444

Wu, Z., Jiang, X., 2023. Extraction of Pine Wilt Disease Regions Using UAV RGB Imagery and Improved Mask R-CNN Models Fused with ConvNeXt. Forests 14, 1672. https://doi.org/10.3390/f14081672

Xu, W., Yang, W., Chen, S., Wu, C., Chen, P., Lan, Y., 2020. Establishing a model to predict the single boll weight of cotton in northern Xinjiang by using high resolution UAV remote sensing data. Computers and Electronics in Agriculture 179, 105762. https://doi.org/10.1016/j.compag.2020.105762

Yang, Y., Xu, Z., Chen, L., Shen, W., Li, H., Zhang, C., Sun, L., Guo, X., Guan, F., 2025. The Aboveground Carbon Stock of Moso Bamboo Forests Is Significantly Reduced byPantana phyllostachysae Chao Stress: Evidence from Multi-source Remote Sensing Imagery. Photogrammetric Engineering & Remote Sensing 91, 213–224. https://doi.org/10.14358/PERS.24-00107R2

Ye, X., Pan, J., Shao, F., Liu, G., Lin, J., Xu, D., Liu, J., 2024. Exploring the potential of visual tracking and counting for trees infected with pine wilt disease based on improved YOLOv5 and StrongSORT algorithm. Computers and Electronics in Agriculture 218, 108671. https://doi.org/10.1016/j.compag.2024.108671

Yu, R., Huo, L., Huang, H., Yuan, Y., Gao, B., Liu, Y., Yu, L., Li, H., Yang, L., Ren, L., Luo, Y., 2022. Frontiers | Early detection of pine wilt disease tree candidates using time-series of spectral signatures. https://doi.org/10.3389/fpls.2022.1000093

Yu, R., Luo, Y., Zhou, Q., Zhang, X., Wu, D., Ren, L., 2021. A machine learning algorithm to detect pine wilt disease using UAV-based hyperspectral imagery and LiDAR data at the tree level. International Journal of Applied Earth Observation and Geoinformation 101, 102363. https://doi.org/10.1016/j.jag.2021.102363

Yuan, Q., Zou, S., Wang, H., Luo, W., Zheng, X., Liu, L., Meng, Z., 2024. A Lightweight Pine Wilt Disease Detection Method Based on Vision Transformer-Enhanced YOLO. Forests 15, 1050. https://doi.org/10.3390/f15061050

Zhang, B., Ye, H., Lu, W., Huang, W., Wu, B., Hao, Z., Sun, H., 2021. A

Spatiotemporal Change Detection Method for Monitoring Pine Wilt Disease in a Complex Landscape Using High-Resolution Remote Sensing Imagery. Remote Sensing 13, 2083. https://doi.org/10.3390/rs13112083

Zhang, N., Zhang, X., Yang, G., Zhu, C., Huo, L., Feng, H., 2018. Assessment of defoliation during the *Dendrolimus tabulaeformis Tsai et Liu* disaster outbreak using UAV-based hyperspectral images. Remote Sensing of Environment 217, 323–339. https://doi.org/10.1016/j.rse.2018.08.024

Zhang, Z., Han, C., Wang, X., Li, H., Li, J., Zeng, J., Sun, S., Wu, W., n.d. Frontiers | Large field-of-view pine wilt disease tree detection based on improved YOLO v4 model with UAV images. https://doi.org/10.3389/fpls.2024.1381367

Zhou, Y., Liu, W., Bi, H., Chen, R., Zong, S., Luo, Y., 2022. A Detection Method for Individual Infected Pine Trees with Pine Wilt Disease Based on Deep Learning. Forests 13, 1880. https://doi.org/10.3390/f13111880

Zhu, X., Wang, R., Shi, W., Liu, X., Ren, Y., Xu, S., Wang, X., 2024. Detection of Pine-Wilt-Disease-Affected Trees Based on Improved YOLO v7. Forests 15, 691. https://doi.org/10.3390/f15040691